%% file: origami-tcas-vt-v29.tex
\documentclass[journal]{IEEEtran}

\newif\ifarxiv
\arxivtrue 

\usepackage{blindtext}
\usepackage{graphicx}

\usepackage[utf8]{inputenc}
\usepackage{textcomp}
\usepackage{gensymb}

\usepackage{pgfplots}
\ifarxiv
	\pgfplotsset{compat=newest}
\else
	\pgfplotsset{compat=1.9}
\fi

\ifCLASSINFOpdf
\else
\fi
\usepackage[cmex10]{amsmath}
\usepackage{algorithmic}

\usepackage{array}
\usepackage{booktabs} 
\usepackage{tabularx} 
\usepackage{threeparttable} 
\usepackage{colortbl}

\usepackage{fixltx2e}

\usepackage{todonotes}
\usepackage{url}

\usepackage{epstopdf}

	\graphicspath{{figs/},{./}}

\begin{document}
\bstctlcite{myBibConfig:BSTcontrol} 
%
\title{Origami: A 803 GOp/s/W \\ Convolutional Network Accelerator}
%
%
%

\author{Lukas~Cavigelli,~\IEEEmembership{Student~Member,~IEEE,}
        and~Luca~Benini,~\IEEEmembership{Fellow,~IEEE}%
\thanks{L. Cavigelli and L. Benini are with the Department
of Electrical Engineering and Information Technology, ETH Zurich, 8092 Zurich, Switzerland. E-mail: \{cavigelli, benini\}@iis.ee.ethz.ch.}%
\thanks{This work was funded by armasuisse Science \& Technology and the ERC MultiTherman project (ERC-AdG-291125).}
\thanks{The authors would like to thank David Gschwend, Christoph Mayer and Samuel Willi for their contributions during design and testing of Origami.}%
\ifarxiv%
	\thanks{}%
\fi%
}%
%
%

\ifarxiv
	\markboth{arXiv preprint}{Cavigelli \MakeLowercase{\textit{et al.}}: Origami}
\fi
%



\maketitle

\begin{abstract}
An ever increasing number of computer vision and image/video processing challenges are being approached using deep convolutional neural networks, obtaining state-of-the-art results in object recognition and detection, semantic segmentation, action recognition, optical flow and superresolution. Hardware acceleration of these algorithms is essential to adopt these improvements in embedded and mobile computer vision systems. 
We present a new architecture, design and implementation as well as the first reported silicon measurements of such an accelerator, outperforming previous work in terms of power-, area- and I/O-efficiency.
The manufactured device provides up to 196\,GOp/s on 3.09\,$\text{mm}^2$ of silicon in UMC 65\,nm technology and can achieve a power efficiency of 803~GOp/s/W. The massively reduced bandwidth requirements make it the first architecture scalable to TOp/s performance. 

\end{abstract}



%
\IEEEpeerreviewmaketitle

\section{Introduction}
\IEEEPARstart{T}{oday} computer vision technologies are used with great success in many application areas, solving real-world problems in entertainment systems, robotics and surveillance~\cite{Porikli2013}. More and more researchers and engineers are tackling action and object recognition problems with the help of brain-inspired algorithms, featuring many stages of feature detectors and classifiers, with lots of parameters that are optimized using the wealth of data that has recently become available. These “deep learning” techniques are achieving record-breaking results on very challenging problems and datasets, outperforming either more mature concepts trying to model the specific problem at hand~\cite{Krizhevsky2012a,Szegedy2014,Sermanet2013,Kaluarachchi2015,Revaud2015} or joining forces with traditional approaches by improving intermediate steps~\cite{Zbontar2015,Taigman2013}. Convolutional Networks (ConvNets) are a prime example of this powerful, yet conceptually simple paradigm \cite{LeCun1989,LeCun2010}. They can be applied to various data sources and perform best when the information is spatially or temporally well-localized, but still has to be seen in a more global context such as in images. 

As a testimony of the success of deep learning approaches, several research programs have been launched, even by major global industrial players (e.g. Facebook, Google, Baidu, Microsoft, IBM), pushing towards deploying services based on  brain-inspired machine learning to their customers within a production environment~\cite{Szegedy2014,Taigman2013,Lin2014}. These companies are mainly interested in running such algorithms on powerful compute clusters in large data centers. 

With the increasing number of imaging devices the importance of digital signal processing in imaging continues to grow. The amount of on- and near-sensor computation is rising to thousands of operations per pixel, requiring powerful energy-efficient digital signal processing solutions, often co-integrated with the imaging circuitry itself to reduce overall system cost and size \cite{Shi2014}. Such embedded vision systems that extract meaning from imaging data are enabled by more and more energy-efficient, low-cost integrated parallel processing engines (multi-core DSPs, GPUs, platform FPGAs). This permits a new generation of distributed computer vision systems, which can bring huge value to a vast range of applications by reducing the costly data transmission, forwarding only the desired information~\cite{Porikli2013,Labovitz2010}. 

Many opportunities for challenging research and innovative applications will pan out from the evolution of advanced embedded video processing and future situational awareness systems. As opposed to conventional visual monitoring systems (CCTVs, IP cameras) that send the video data to a data center to be stored and processed, embedded smart cameras process the image data directly on board. This can significantly reduce the amount of data to be transmitted and the required human intervention – the sources of the two most expensive aspects of video surveillance \cite{Bobda2014}. Embedding convolutional network classifiers in distributed computer vision systems, seems a natural direction of evolution, However, deep neural networks are commonly known for their demand of computing power, making it challenging to bring this computational load within the power envelope of embedded systems -- in fact, most state-of-the-art neural networks are currently not only trained, but also evaluated on workstations with powerful GPUs to achieve reasonable performance.

 Nevertheless, there is strong demand for mobile vision solutions ranging from object recognition to advanced human-machine interfaces and augmented reality. The market size is estimated to grow to many billions of dollars over the next few years with an annual growth rate of more than 13\% \cite{MarketsAndMarkets2014}. This has prompted many new commercial solutions to become available recently, specifically targeting the mobile sector \cite{Movidius2014,Campbell2015,Mobileye2011}. 

In this paper we present: 
\begin{itemize}
	\item The \emph{architecture} of a novel convolutional network accelerator, which is scalable to TOP/s performance while remaining area- and energy-efficient and keeping I/O throughput within the limits of economical packages and low power budgets. This extends our work in \cite{Cavigelli2015a}. 
	\item An \emph{implementation} of this architecture with optimized precision using fixed-point evaluations constrained for an accelerator-sized ASIC. 
	\item \emph{Silicon measurements} of the taped-out ASIC, providing experimental characterization of the silicon. 
	\item A thorough \emph{comparison} to and discussion of previous work.
\end{itemize}

\paragraph*{Organization of the paper}
Section~\ref{sec:convNets} shortly introduces convolutional networks and highlights the need for acceleration. Previous work is investigated in Section~\ref{sec:prevWork}, discussing available software, FPGA and ASIC implementations and explaining the selection of our design objectives. In Section~\ref{sec:arch} we present our architecture and its properties. The implementation aspects are shown in Section~\ref{sec:impl}. We present our results in Section~\ref{sec:results} and discuss and compare them in Section~\ref{lbl:discussion}. We conclude the paper in Section~\ref{sec:conclusions}.

\section{Convolutional Networks}
\label{sec:convNets}
Most convolutional networks (ConvNets) are built from the same basic building blocks: convolution layers, activation layers and pooling layers. One sequence of convolution, activation and pooling is considered a \emph{stage}, and modern, deep networks often consist of multiple stages. The convolutional network itself is used as a feature extractor, transforming raw data into a higher-dimensional, more meaningful representation. ConvNets particularly preserve locality through their limited filter size, which makes them very suitable for visual data (e.g., in a street scene the pixels in the top left corner contain little information on what is going on in the bottom right corner of an image, but if there are pixels showing the sky all around some segment of the image, this segment is certainly not a car). The feature extraction is then followed by a classifier, such as a normal neural network or a support vector machine. 

A stage of a ConvNet can be captured mathematically as 
\begin{align}
    \mathbf{y}^ {(\ell)} &= \text{conv}(\mathbf{x}^{(\ell)},\mathbf{k}^{(\ell)})+\mathbf{b}^{(\ell)},\\
    \mathbf{x}^ {(\ell+1)}&=\text{pool}(\text{act}(\mathbf{y}^ {(\ell)})), 
\end{align}
where $\ell=1\dots 3$ indexes the stages and where we start with $\mathbf{x}^{(1)}$ being the input image. 
The key operation on which we focus is the convolution, which expands to 
\begin{align}
    &y^{(\ell)}_o(j,i) = b^{(\ell)}_o + \sum_{c\in\mathcal{C}_{in}^{(\ell)}}  \sum_{(b, a)\in\mathcal{S}^{(\ell)}} k^{(\ell)}_{o,c}(b, a) x^{(\ell)}_c(j - b, i - a),
\end{align}
where $o$ indexes the output channels $\mathcal{C}_{out}^{(\ell)}$ and $c$ indexes the input channels $\mathcal{C}_{in}^{(\ell)}$. The pixel is identified by the tuple $(j,i)$ and $\mathcal{S}_k$ denotes the support of the filters. In recently published networks \cite{Dong2014,Farabet2013,Szegedy2014}, the pooling operation determines the maximum in a small neighborhood for each channel, often on $2 \times 2$ areas and with a stride of $2 \times 2$. $\mathbf{x} = \text{pool}_{\text{max},2\times 2}(\mathbf{v})$: 
\begin{alignat}{2}
	x_o(j,i) 	= \text{max}(	&v_o(2j,2i), &&v_o(2j,2i+1),\nonumber\\
						&v_o(2j+1,2i),\,&&v_o(2j+1,2i+1)) \label{eqn:maxPool2}
\end{alignat}
The activation function is applied point-wise for every pixel and every channel. A currently popular choice is the rectified linear unit (ReLU) \cite{Krizhevsky2012a,Sermanet2013,Kaluarachchi2015}, which designates the function $x \mapsto \max(0,x)$. The activation function introduces non-linearity into neural networks, giving them the potential to be more powerful than linear methods. Typical filter sizes range from $5 \times 5$ to $9 \times 9$, sometimes even $11 \times 11$ \cite{Krizhevsky2012a,Sermanet2013,Farabet2013}. 
\begin{align}
	\mathbf{v} = \text{act}_\text{ReLU}(\mathbf{y}), \quad v_o(j,i) = \text{max}(y_o(j,i),0)
\end{align}

The feature extractor with the convolutional layers is usually followed by a classification step with fully-connected neural network layers interspersed with activation functions, reducing the dimensionality from several hundred or even thousands down to the number of classes. In case of scene labeling these fully-connected layers are just applied on a per-pixel basis with inputs being the values of all the channels at any given pixel pixel \cite{Long2015}.

\subsection{Measuring Computational Complexity}
Convolutional networks and deep neural networks in general are advancing into more and more domains of computer vision and are becoming increasingly more accurate in their traditional application area of object recognition and detection. ConvNets are now able to compute highly accurate optical flow \cite{Kaluarachchi2015,Revaud2015,Weinzaepfel2013}, super-resolution \cite{Dong2014} and more. The newer networks are usually deeper and require more computational effort, and those for the newly tapped topics have already been very deep from the beginning. Research is done on various platforms and computing devices are evolving rapidly, making time measurements meaningless. The deep learning community has thus started to measure the complexity of deep learning networks in a way that is more independent of the underlying computing platform, counting the additions and multiplications of the synapses of these networks. For a convolutional layer with $n_{in}$ input feature maps of size $h_{in}\times w_{in}$, a filter kernel size of $h_k \times w_k$, and $n_{out}$ output feature maps, this number amounts to 
\begin{align}
	2n_{out}n_{in}h_k w_k (h_{in} - h_k + 1)(w_{in} - w_k + 1),
\end{align} 
where $n_{out}$ is the number of output channels $|\mathcal{C}_{out}|$, $n_{in}$ is the number of $|\mathcal{C}_{in}|$, $h_{in} \times w_{in}$ is the size of the image and $h_k \times w_k$ is the size of the filter in spatial domain. The factor of two is because the multiplications and additions are counted as separate operations in this measure, which is the most common in neural network literature \cite{Farabet2011,Pham2012,Chen2014,Cavigelli2015}. 

However, this way of measuring complexity still does not allow to perfectly determine how a network performs on different platforms. Accelerators might need to be initialized or have to suspend computation to load new filter values, often performing better for some artificially large or small problems. For this reason we distinguish between the throughput obtained with a real network (\emph{actual throughput} or just \emph{throughput}), measurements obtained with a synthetic benchmark optimized to squeeze out the largest possible value (\emph{peak throughput}), and the maximum throughput of the computation units without caring for bandwidth limits often stated in the device specifications of non-specialized processors (\emph{theoretical throughput}). 

Software and hardware implementations alike often come with a throughput dependent on the actual size of the convolutional layer. 
While we make sure our chip can run a large range of ConvNets efficiently, we use the one presented in \cite{Cavigelli2015} as a reference for performance evaluation. It has three stages and we assume input images of size $240 \times 320$. The resulting sizes and complexities of the individual layers are summarized in Table~\ref{tbl:refConvNetParams} and use a filter of size $7 \times 7$ for all of them. The total number of operations required is 7.57\,GOp/frame. To give an idea of the complexity of more well-known ConvNets, we have listed some of them in Table~\ref{tbl:knownConvNetComplexities}. If we take an existing system like the NeuFlow SoC~\cite{Pham2012} which is able to operate at 490\,GOp/s/W, we can see that very high quality, dense optical flow on $384\times 512$ video can be computed with 25 frame/s at a power of just around 3.5\,W if we could scale up the architecture. We can also see that an optimized implementation on a high-end GPU can run at around 27 frame/s. 

%

\begin{table}
	\caption{Parameters of the Three Stages of Our Reference Scene Labeling Convolutional Network. }
	\label{tbl:refConvNetParams}
	\centering
	\small
	\begin{tabular}{lrrrr}
		\toprule
		& Stage 1 & Stage 2 & Stage 3 & Classif. \\ \midrule
		Input size & 240$\times$320 & 117$\times$157 & 55$\times$75 & 49$\times$69 \\
		\# Input ch. &	3 &	16 &	64 & 256\\
		\# Output ch. &	16 &	64 &	256 & 8 \\
		\# Operations &	346\,MOp &	1682\,MOp &	5428\,MOp & 115\,MOp \\
		\# Filter val. &	2.4k &	50k &	803k & 17k \\
		\bottomrule
	\end{tabular}
\end{table}

\subsection{Computational Effort}
Because convolutional networks can be evaluated significantly faster than traditional approaches of comparable accuracy (e.g. graphical models), they are approaching an area where real-time applications become feasible on workstations with one, or more often, several GPUs. However, most application areas require a complete solution to fit within the power envelope of an embedded systems or even a mobile device. Taking the aforementioned scene labeling ConvNet as an examples, its usage in a real-time setting at 25~frame/s amounts to 189~GOp/s, which is out of the scope of even the most recent commercially available mobile processors \cite{Cavigelli2015}. 

\begin{figure}
	\includegraphics[width=\columnwidth]{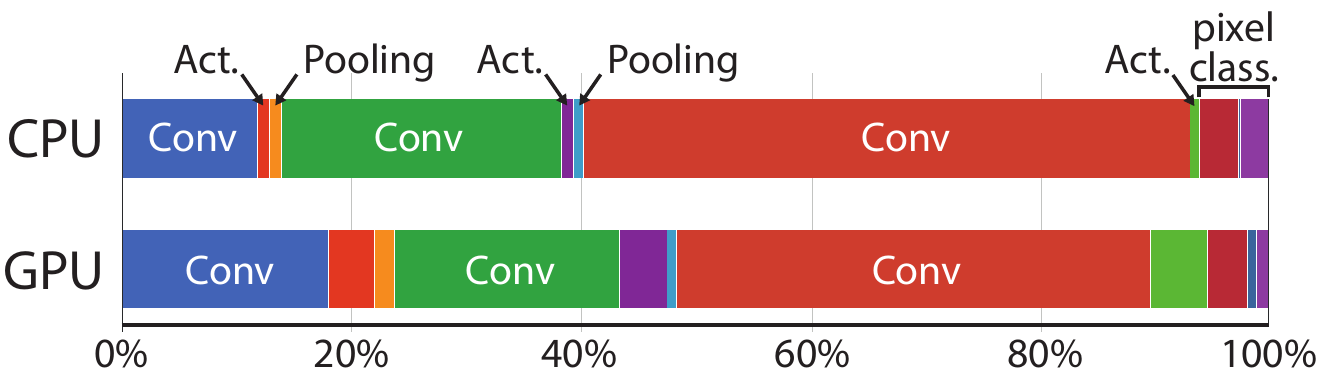}
	\caption{Computation time spent in different stages of our reference scene labeling convolutional network~\cite{Cavigelli2015}.}
	\label{fig:timeSplit}
\end{figure}

For a subject area changing as rapidly as deep learning, the long-term usability is an important objective when thinking about hardware acceleration of the building blocks of such systems. While the structure of the networks is changing from application to application and from year to year, and better activation and pooling operations are continuously being published, there is a commonality between all these ConvNets: the convolutional layer. It has been around since the early 90s and has not changed since~\cite{LeCun1989,Sermanet2013,Szegedy2014}. Fortunately, this key element is also the computation-intensive part for well-optimized software implementations (approx. 89\% of the total computation time on the CPU, or 79\% on the GPU) as shown in Figure~\ref{fig:timeSplit}. The time for activation and pooling is negligible as well as the computation time for the pixel-wise classification with fully-connected layers. 

\begin{table}
	\caption{Number of Operations Required to Evaluate Well-Known Convolutional Networks.}
	\label{tbl:knownConvNetComplexities}
	\centering
	\begin{tabular}{lllr}
	\toprule
	name & type & challenge/dataset & \# GOp \\ \midrule
	\cite{Cavigelli2015} SS 320$\times$240 & scene labeling & stanford backgr., 74.8\% & 7.57 \\
	\cite{Cavigelli2015} SS full-HD & scene labeling & stanford backgr., 74.8\% & 259.5 \\
	\cite{Cavigelli2015} MS 320$\times$240 & scene labeling & stanford backgr., 80.6\% & 16.1 \\
	AlexNet & image recog. & imagenet/ILSVRC 2012 & 1.7 \\ 
	OverFeat fast & image recog. & imagenet/ILSVRC 2013 & 5.6 \\
	OverFeat accurate & image recog. & imagenet/ILSVRC 2013 & 10.7 \\
	GoogLeNet & image recog. & imagenet/ILSVRC 2014 & 3.6 \\
	VGG Oxfordnet A & image recog. & imagenet/ILSVRC 2014 & 15.2 \\ 
	FlowNetS(-ft) & optical flow & synthetic \& KITTI, Sintel & 68.9 \\
	\bottomrule	
	\end{tabular}
\end{table}

\section{Previous Work}
\label{sec:prevWork}

Convolutional Networks have been achieving amazing results lately, even outperforming humans in image recognition on large and complex datasets such as Imagenet. The top performers have achieved a top-5 error rate (actual class in top 5 proposals predicted by the network) of only 6.67\% (GoogLeNet \cite{Szegedy2014}) and 7.32\% (VGG Oxfordnet \cite{Simonyan2015}) at the ILSVRC 2014 competition \cite{Russakovsky2014}. The best performance of a single human so far is 5.1\% on this dataset and has been exceeded since the last large image recognition competition \cite{He2015}. Also in other subjects such as face recognition \cite{Taigman2013}, ConvNets are exceeding human performance. We have listed the required number of operations to evaluate some of these networks in Table~\ref{tbl:knownConvNetComplexities}. 

In the remainder of this section, we will focus on existing implementations to evaluate such ConvNets. We compare software implementations running on desktop workstations with CPUs and GPUs, but also DSP-based works to existing FPGA and ASIC implementations. In Section~\ref{sec:generalNNacc} we discuss why many such accelerators are not suitable to evaluate networks of this size and conclude the investigation into previous work by discussing the limitation of existing hardware architectures in Section~\ref{sec:prevWorkDiscussion}.

\subsection{Software Implementations (CPU, GPU, DSP)}
Acceleration of convolutional neural networks has been discussed in many papers. There are very fast and user-friendly frameworks publicly available such as Torch~\cite{Collobert2011}, Caffe~\cite{Jia2013}, Nvidia's cuDNN~\cite{Chetlur2014} and Nervana~Systems' neon~\cite{nervana2015}, and GPU-accelerated training and evaluation are the commonly way of working with ConvNets. 

These and other optimized implementations can be used to obtain a performance and power efficiency baseline on desktop workstations and CUDA-compatible embedded processors, such as the Tegra~K1. On a GTX780 desktop GPU, the performance can reach up to 3059\,GOp/s for some special problems and about 1800\,GOp/s on meaningful ConvNets. On the Tegra~K1 up to 96\,GOp/s can be achieved, with 76\,GOp/s being achieved with an actual ConvNet. On both platforms an energy-efficiency of about 7\,GOp/s/W considering the power of the entire platform and 14.4\,GOp/s/W with differential power measurements can be obtained \cite{Cavigelli2015}. Except for this evaluation the focus is usually on training speed, where multiple images are processed together in batches to attain higher performance (e.g. using the loaded filter values for multiple images). Batch processing is not suitable for real-time applications, since it introduces a delay of many frames. 

A comparison of the throughput of many optimized software implementations for GPUs based on several well-known ConvNets is provided in \cite{ChintalaBenchm}. The list is lead by an implementation by Nervana Systems of which details on how it works are not known publicly. They confirm that it is based on maxDNN~\cite{Lavin2015}, which started from an optimized matrix-matrix multiplication, adapted for convolutional layers and with fine-tuned assembly code. Their implementation is tightly followed by Nvidia's cuDNN library~\cite{Chetlur2014}. The edge of these two implementations over others originates from using half-precision floating point representations instead of single-precision for storage in memory, thus reducing the required memory bandwidth, which is the currently limiting factor. New GPU-based platforms such as the Nvidia Tegra~X1 are now supporting half-precision computation \cite{Nvidia2015}, which can be used to save power or provide further speedup, but no thorough investigations have been published on this. 
More computer vision silicon has been presented recently with the Movidius Myriad~2 device \cite{Movidius2014} which has been used in Google~Tango, and the Mobileye EyeQ3 platform, but no benchmarking results regarding ConvNets are available yet.

A different approach to increase throughput is through the use of the Fourier transform, diagonalizing the convolution operation. While this has a positive effect for kernels larger than $9\times 9$, the bandwidth problem generally becomes much worse and the already considerable memory requirements are boosted further, since the filters have to be padded to the input image size~\cite{Mathieu2013,Cavigelli2015}. 

However optimized the software running on such platforms, it will always be constrained by the underlying architecture: the arithmetic precision cannot be adapted to the needs of the computation, caches are used instead of optimized on-chip buffers, instructions have to be loaded and decoded. This pushes the need for specialized architectures to achieve high power- and area-efficiency.

\subsection{FPGA Implementations}
Embeddability and energy-efficiency is a major concern regarding commercialization of ConvNet-based computer vision systems and has hence prompted many researchers to approach this issue using FPGA implementations. Arguably the most popular architecture is the one which started as CNP~\cite{Farabet2009} and was further improved and renamed to NeuFlow~\cite{Farabet2011,Pham2012} and later on to nn-X~\cite{Gokhale2014}. 

Published in 2009, CNP was the first ConvNet specific FPGA implementation and achieved 12~GOp/s at 15~W on a Spartan 3A DSP 3400 FPGA using 18~bit fixed-point arithmetic for the multiplications. Its architecture was designed to be self-contained, allowing it to execute the operations for all common ConvNet layers, and coming with a soft CPU to control the overall program flow. It also features a compiler, converting network implementations with Torch directly to CNP instructions.   

The CNPs architecture does not allow easy scaling of its performance, prompting the follow-up work NeuFlow which uses multiple CNP convolution engines, an interconnect, and a smart DMA controller. The data flow between the processing tiles can be be rerouted at runtime. The work published in 2011 features a Virtex 6 VLX240T to achieve 147\,GOp/s at 11\,W using 16\,bit fixed-point arithmetic. 

To make use of the newly available platform ICs, NeuFlow was ported to a Zynq~XC7Z045 in 2014, further improved by making use of the hard-wired ARM cores, and renamed to nn-X. It further increases the throughput to about 200\,GOp/s at 4\,W (FPGA, memory and host) and uses $4 \times 950$\,MB/s full-duplex memory interfaces. 

Only few alternatives to CNP/NeuFlow/nn-X exist. The two most relevant are a ConvNet accelerator based on Microsoft's Catapult platform in~\cite{Ovtcharov2015} with very little known details and a HLS-based implementation \cite{Zhang2015} with a performance and energy efficiency inferior to nn-X. 


\subsection{ASIC Implementations}
The NeuFlow architecture was implemented as an ASIC in 2012 on $12.5\,\text{mm}^2$ of silicon for the IBM 45nm SOI process. The results based on post-layout simulations were published in~\cite{Pham2012}, featuring a performance of about 300\,GOp/s at 0.6\,W operating at 400\,MHz with an external memory bandwidth of $4 \times 1.6$\,GB/s full-duplex. 

To explore the possibilities in terms of energy efficiency, a convolution accelerator suitable for small ConvNets was implemented in ST 28nm FDSOI technology~\cite{Conti}. They achieve 37\,GOp/s with 206\,GOp/s/W at 0.8\,V and 1.39\,GOp/s with 1375\,GOp/s/W at 0.4\,V during simulation (pre-silicon) with the same implementation, using aggressive voltage scaling combined with reverse body biasing available with FDSOI technology. 

Further interesting aspects are highlighted in ShiDianNao~\cite{Du2015,Chen2015}, which evolved from DianNao~\cite{Chen2014}. The original DianNao was tailored to fully-connected layers, but was also able to evaluate convolutional layers. However, its buffering strategy was not making use of the 2D structure of the computational problem at hand. This was improved in ShiDianNao. Nevertheless, its performance strongly depends on the size of the convolutional layer to be computed, only unfolding its performance for tiny feature maps and networks. They achieve a peak performance of 128\,GOp/s with 320\,mW on a core-only area of $1.3\text{mm}^2$ in a TSMC 65\,nm post-layout evaluation. 

Another way to approach the problem at hand is to look at general convolution accelerators, such as the ConvEngine~\cite{Qadeer2013} which particularly targets 1D and 2D convolutions common in computer vision applications. It comes with an array of 64 10-bit ALUs and input and output buffers optimized for the task at hand. Based on synthesis results, they achieve a core-only power efficiency of 409~GOp/s/W. 

In the last few months we have seen a wave of vision DSP IP cores and SoCs becoming commercially available: CEVA-XM4, Synopsys DesignWare EV5x, Cadence Tensilica Vision P5. They are all targeted at general vision applications and not specifically tailored to ConvNets. They are processor-based and use vector engines or many small specialized processing units. 
Many of the mentioned IP blocks have never been implemented in silicon, and their architecture is kept confidential and has not been peer reviewed, making a quantitative comparison impossible. 
However, as they use instruction-based processing, an energy efficiency gap of $10\times$ or more with respect to specialized ASICs can be expected. 

\subsection{General Neural Network Accelerators}
\label{sec:generalNNacc}
Besides the aforementioned efforts, there are many accelerators which are targeted at accelerating non-convolutional neural networks. One such accelerator is the K-Brain~\cite{Park2015a}, which was evaluated to achieve an outstanding power efficiency of 1.93\,TOp/s/W in 65\,nm technology. It comes with 216\,KB of SRAM to store the weights and the dataset. For most applications this is by far insufficient (GoogLeNet~\cite{Szegedy2014}: 6.8M, VGG-Oxfordnet~\cite{Simonyan2015}: 133M parameters) and the presented architectures do not scale to larger networks, requiring excessive amounts of on-chip memory \cite{Park2015a,Yu2015,Moreau2015,Zhang2015a}. Other neural network accelerators are targeted at more experimental concepts like spiking neural networks, where thorough performance evaluations are still missing \cite{Akopyan2015}.

\subsection{Discussion}
\label{sec:prevWorkDiscussion}
Recent work on hardware accelerators for ConvNets shows that highly energy-efficient implementations are feasible, significantly improving over software implementations. 

However, existing architectures are not scalable to higher performance applications as a consequence of their  need for a very wide memory interface. This manifests itself with the 299~I/O pins required to achieve 320\,GOp/s using NeuFlow~\cite{Pham2012}. For many interesting applications, much higher throughput is needed, e.g. scene labeling of full-HD frames requires 5190\,GOp/s to process 20\,frame/s, and the trend clearly points towards even more complex ConvNets. To underline the need for better options, we want to emphasize that linearly scaling NeuFlow would require almost 5000~I/O pins or 110\,GB/s full-duplex memory bandwidth. This issue is currently common to all related work, as long as the target application is not limited to tiny networks which allow caching of the entire data to be processed. 

This work particularly focuses on this issue, reducing the memory bandwidth required to achieve a high computational throughput without using very large on-chip memories to store the filters and intermediate results. 
For state-of-the-art networks storing the learned parameters on-chip is not feasible with GoogLeNet requiring 6.8\,M and VGG~Oxfordnet 135\,M parameters. The aforementioned scene labeling ConvNet required 872\,k parameters, of which 855\,k parameters are filter weights for the convolutional layers. Some experiments have been done on the required word width \cite{Courbariaux2015,Gupta2015,Vanhoucke2011} and compression \cite{Soulie2015}, but validated only on very small datasets (MNIST, CIFAR-10). 

\section{Architecture}
\label{sec:arch}
\begin{figure}
	\includegraphics[width=\columnwidth]{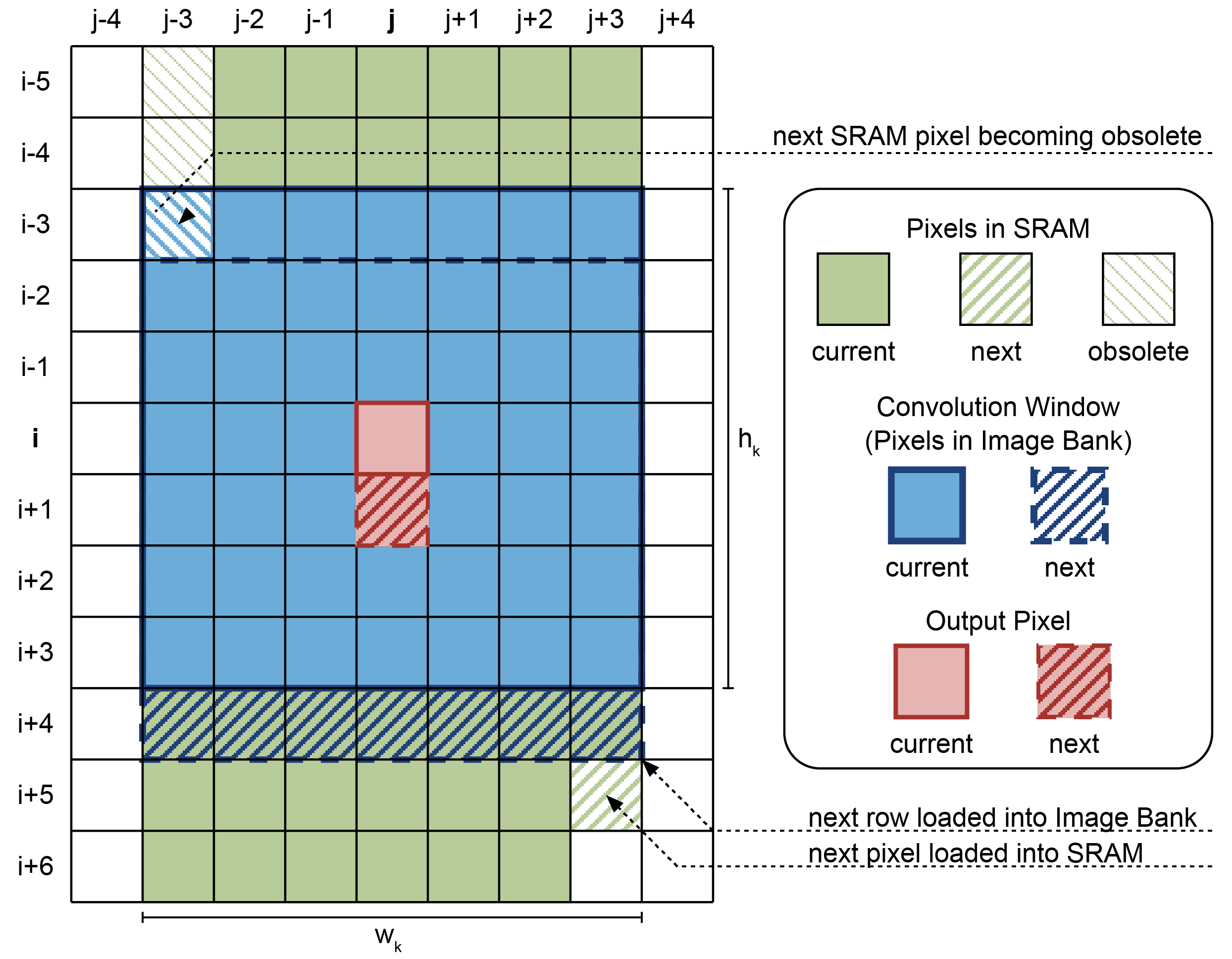}
	\caption{Data stored in the image bank and the image window SRAM per input channel.}
	\label{fig:memWindow}
\end{figure}
\begin{figure*}
		\includegraphics[width=\textwidth]{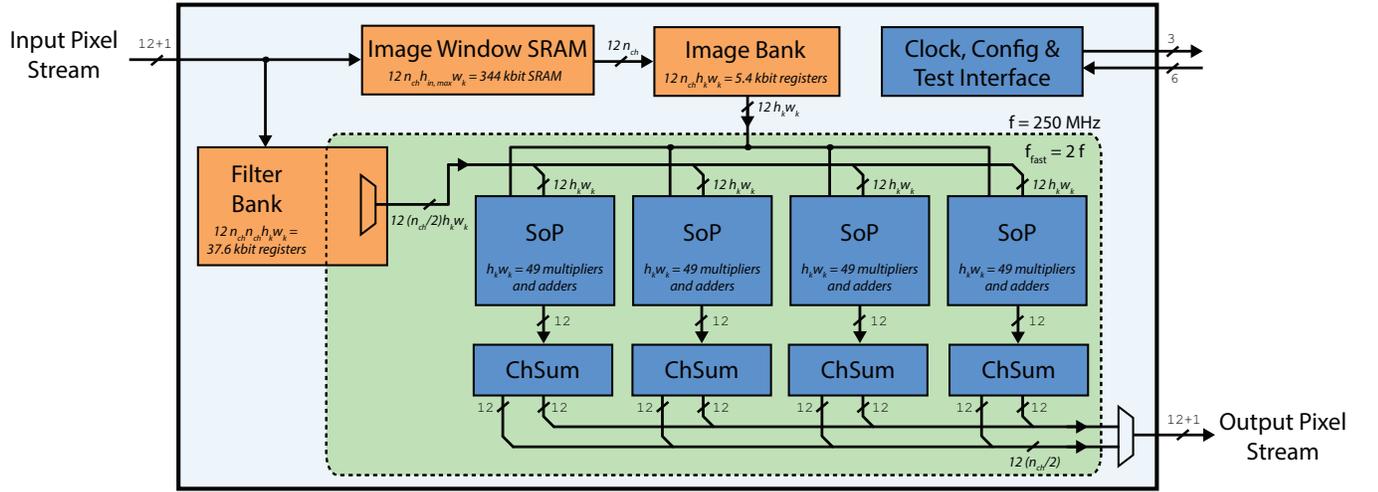}
	\caption{Top-level block diagram of the proposed architecture for the chosen implementation parameters.}
	\label{fig:topLevelBlockDiag}
\end{figure*}
\begin{figure}
	\centering
	\def\svgwidth{\columnwidth}
	\ifarxiv
		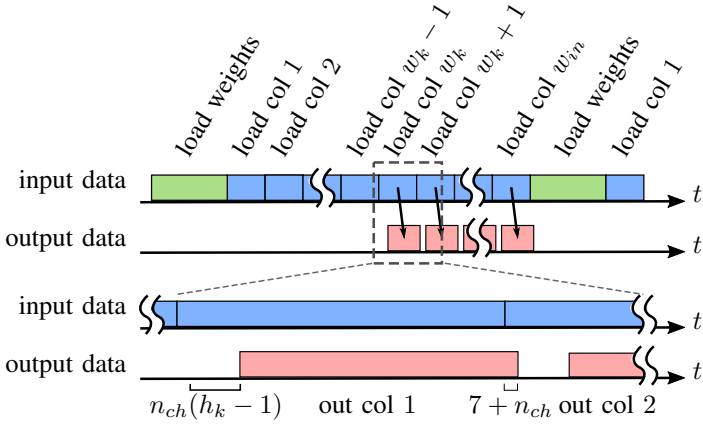
	\fi
	\caption{Time diagram of the input and output data transfers. Internal activity is strongly related up to small delays (blue = load col = image window SRAM r/w active = image bank write active; red = output = ChSum active = SoP active = image bank reading = filter bank reading; green = load weights = filter bank shifting/writing)}
	\label{fig:timeline}
\end{figure}

In this section we first present the concept of operation of our architecture in a simple configuration. We then explain some changes which make it more suitable for an area-efficient implementation. We proceed by looking into possible inefficiencies when processing ConvNet data. We conclude this section by presenting a system architecture suitable to embed Origami in a SoC or a FPGA-based system. 

\subsection{Concept of Operation}
A top-level diagram of the architecture is shown in Figure~\ref{fig:topLevelBlockDiag}. It shows two different clock areas which are explained later on. The concept of operation for this architecture first assumes a single clock for the entire circuit for simplicity. In Figure~\ref{fig:timeline} we show a timeline with the input and output utilization. Note that the utilization of internal blocks corresponds to these utilizations in a very direct way up to a short delay. 

The input data (image with many channels) are fed in stripes of configurable height into the circuit and stored in a SRAM, which keeps a spatial window of the input image data. The data is the loaded into the image bank, where a smaller window of the size of the filter kernel is kept in registers and moved down on the image stripe before jumping to the next column. This register-based memory provides the input for the sum-of-product (SoP) units, where the inner products of the individual filter kernels are computed. Each SoP unit is fed the same image channel, but different filters, such that each SoP computes the partial sum for a different output channel. The circuit iterates over the channels of the input image while the partial sums are accumulated in the channel summer (ChSum) unit to compute the complete result, which is then transmitted out of the circuit. 

For our architecture we tile the convolutional layer into blocks with a fixed number of input and output channels $n_{ch}$. We perform $2 n_{ch}^2 h_k w_k$ operations every $n_{ch}$ clock cycles, while transmitting and receiving $n_{ch}$ values instead of $n_{ch}^2$. This is different from all previous work, and improves the throughput per bandwidth by a factor of $n_{ch}$. The architecture can also be formulated for non-equal block size for the input and output channels, but there is no advantage doing so, thus we keep this constraint for simplicity of notation. 
We proceed by presenting the individual blocks of the architecture in more detail. 



\subsubsection{Image Window SRAM and Image Bank}
The image window SRAM and the image bank are in charge of storing new received image data and providing the SoP units with the image patch required for every computation cycle. To minimize size, we want to keep the image bank as small as possible, while not requiring an excessive data rate from the SRAM. The size of the image bank was chosen as $n_{ch} h_k w_k$. In every cycle a new row of $w_k$ of the current input channel elements is loaded from the image window SRAM and shifted into the image bank. The situation is illustrated in Figure~\ref{fig:memWindow} for an individual channel.

In order for the SRAM to be able to provide this minimum amount of data, it needs to store a $w_k$ element wide window for all $n_{ch}$ channels and have selectable height $h_{in}\leq h_{(in,max)}$. A large image has to be fed into the circuit in stripes of maximum height $h_{in,max}$ with an overlap of $h_k-1$ pixels. The overlap is because an evaluation of the kernel will need a surrounding of $(h_k-1)/2$ pixel in height and $(w_k/1)/2$ pixel in width. When the image bank reaches the bottom of the image window stored in SRAM, it jumps back to the top, but shifted one pixel to the right. This introduces a delay of $n_{ch}(h_k-1)$ cycles, during which the rest of the circuit is idling. This delay is not only due to the loading of the new values for the image bank, but also to receive the new pixels for the image window SRAM through the external I/O. Choosing $h_{(in,max)}$ is thus mostly a trade-off between throughput and area. The performance penalty on the overall circuit is about a factor of $(h_k - 1)/h_{(in,max)}$. The same behavior can be observed at the beginning of the horizontal stripe. During the first $n_{ch} h_{in} (w_k-1)$ cycles the processing units are idling.

\subsubsection{Filter Bank}
The filter bank stores all the weights of the filters, these are $n_{ch} n_{ch} h_k w_k$ values. In configuration mode the filter values are shifted into these registers which are clocked with at the lower frequency $f$. 
In normal operation, the entire filter bank is read-only. In each cycle all the filter values supplied to the SoP have to be changed, this means that $n_{ch} h_k w_k$ filter values are read per cycle. Because so many filters have to be read in parallel and they change so frequently, it is not possible to keep them in a SRAM. Instead, it is implemented with registers and a multiplexer capable of multiplexing selecting one of $n_{ch}$ sets of $n_{ch} h_k w_k$ weights. 

The size of the filter bank depends quadratically on the number of channels processed, which results in a trade-off between area and I/O bandwidth efficiency. When doubling the I/O efficiency (doubling $n_{ch}$, doubling I/O bandwidth, quadrupling the number of operations per data word), the storage requirements for the filter bank are quadrupled. 

Global memory structures which have to provide lots of data at a high speed are often problematic during back end design. It is thus important to highlight that while this filter bank can be seen as such a global memory structure, but is actually local: Each SoP unit only needs to access the filters of the output channel it processes, and no other SoP unit accesses these filters. 

\subsubsection{Sum-of-Products Units}
A SoP unit calculates the inner product between an image patch and a filter kernel. It is built from $h_k w_k$ multipliers and $h_k w_k -1$ adders arranged in a tree. Mathematically the output of a SoP unit is described as $\sum_{(\Delta j,\Delta i)\in\mathcal{S}_{k}}k_{o,c}(\Delta j, \Delta i) x_c(j-\Delta j, i-\Delta i)$.

While previous steps have only loaded and stored data, we here perform a lot of arithmetic operations, which raises the question of numerical precision. A fixed-point analysis to select the word-width is shown in Section~\ref{sec:fixedPoint}. In terms of architecture, the word width $v$ is doubled by the multiplier, and the adder tree further adds $\log_2(h_k w_k)$ bits. We truncate the result to the original word width with the same underlying fixed-point representation. This truncation also reduces the accuracy with which the adder tree and the multipliers have to be implemented. The idea of using the same fixed-point representation for the input and output is motivated by the fact that there are multiple convolutional layers and each output will also serve again as an input. 

\subsubsection{Channel Summer Units}
Each ChSum unit sums up the inner products it receives from the SoP unit it is connected to, reducing the amount of data to be transmitted out of the circuit by a factor of $1/n_{ch}$ over the naive way of transmitting the individual convolution results. The SoP units are built to be able to perform this accumulation while still storing the old total results, which are one-by-one transmitted out of the circuit while the next computations are already running. The ChSum units also perform their calculations at full precision and the results are truncated to the original fixed-point format.

\subsection{Optimizing for Area Efficiency}
To achieve a high area efficiency, it is essential that large logic blocks are operated at a high frequency. We can pipeline the multipliers and the adder tree inside the SoP units to achieve the desired clock frequency. The streaming nature of the overall architecture makes it very simple to vary this without drawbacks as encountered with closed-loop architectures. 

The limiting factor for the overall clock frequency is the SRAM keeping the image window, which comes with a fixed delay and a minimum clock period, and the speed of CMOS I/O pads. Because the SRAM's maximum frequency is much lower than the one of the computation-density- and power-optimized SoP units, we have chosen to have them running at twice the frequency. This way each unit calculates two inner products until the image bank changes the channel of the input image or takes a step forward. This makes each SoP unit responsible for two output channels. While there is little change to the image bank, the values taken from the filter bank have to be switched at the faster frequency as well. Additionally, the ChSum units have to be adapted to alternatingly accumulate the inner products of the two different output channels. 

The changes induced to the filter bank reduce the number of filter values to be read to $n_{ch}h_kw_k/2$ per cycle, however at twice the clock rate. The adapter filter bank has to be able to read one of $2n_{ch}$ sets of $n_{ch} h_k w_k/2$ weights each at $f_{fast}=2f$. 

\subsection{Throughput}

\begin{table}
	\caption{Throughput and Efficiency for the Individual Stages of our Reference Convolutional Network for 320$\times$240 Input Images.}
	\label{tbl:throughputEfficiency}
	\centering
	\small
	\begin{tabularx}{\columnwidth}{XXXX}
		\toprule
		stage & Stage 1 & Stage 2 & Stage 3 \\ 
		\# channels & (3$\rightarrow$16) & (16$\rightarrow$64) & (64$\rightarrow$256)\\ \midrule
		$\eta_\text{chIdle}$ & 0.38 & 1.00 & 1.00 \\
		$\eta_\text{filterLoad}$ & 0.99 & 0.98 & 0.91 \\
		$\eta_\text{border}$ & 0.96 & 0.91 & 0.82 \\
		$\eta$ & 0.36 & 0.89 & 0.75 \\ \midrule
		throughput & 71 GOp/s & 174 GOp/s & 147 GOp/s \\
		\# operations & 0.35\;GOp	 & 1.68\;GOp & 5.43\;GOp\\
		run time & 4.93\;ms & 9.65\;ms & 36.94\;ms \\ \midrule
		\multicolumn{4}{c}{Average throughput: 145\;GOp/s $\rightarrow$ 19.4\;frame/s @ 320$\times$240} \\
		\bottomrule
	\end{tabularx}
\end{table}

The peak throughput of this architecture is given by
\begin{align*}
	2 n_\text{SoP} h_k w_k f_{fast} = 2 n_{ch} h_k w_k f
\end{align*} 
operations per second. Looking at the SoP units, they can each calculate $w_k \times h_k$ multiplications and additions per cycle. As mentioned before, the clock is running at twice the speed ($f_{fast} = 2f$) to maximize area efficiency by using only $n_\text{SoP}=n_{ch}/2$ SoP units. All the other blocks of the circuit are designed to be able to sustain this maximum throughput. Nevertheless, several aspects may cause these core operation units to stall. We discuss the aspects in the following paragraphs.


\subsubsection{Border Effects}
At the borders of an image no valid convolution results can be calculated, so the core has to wait for the necessary data to be transferred to the device. These waiting periods occur at the beginning of a new image while $w_k-1$ columns are preloaded, and at the beginning of each new column while $h_k-1$ pixels are loaded in $n_{ch} (h_k - 1)$ cycles. The effective throughput thus depends on the size of the image: 
\begin{align*}
	\eta_\text{border}=(h_{in}-h_k+1)(w_{in}-w_k+1)/(h_{in} w_{in} ).
\end{align*}
The maximum $h_{in}$ is limited to some $h_{in,max}$ depending on the size of the image window SRAM. It is feasible to choose $h_{in}$ large enough to process reasonably sized image, otherwise the image has to be tiled into multiple smaller horizontal stripes with an overlap of $h_k -1$ rows with the corresponding additional efficiency loss.  

Assuming $h_{in,max}$ is large enough and considering our reference network, this factor is 0.96, 0.91 and 0.82 for stages~$1,\dots,3$, respectively in case of a $240 \times 320$~pixel input image. For larger images this is significantly improved, e.g. for a $480 \times 640$~image the Stage~3 will get an efficiency factor of 0.91. However, the height of the input image is limited to 512~pixel due to the memory size of the image bank.

\subsubsection{Filter Loading}
Before the image transmission can start, the filters have to be loaded through the same bus used to transmit the image data. This causes a loss of a few more cycles. Instead of just the $n_{ch}h_{in} w_{in}$ input data values, an additional $n_{ch}^2 h_k w_k$ words with the filter weights have to be transferred. This results in an additional efficiency loss by a factor of
\begin{align*}
	\eta_\text{filterLoad} =  \frac{n_{ch} h_{in} w_{in}}{n_{ch} n_{ch} h_k w_k + n_{ch} h_{in} w_{in}}.
\end{align*}
If we choose $n{ch}=8$, this evaluates to 0.99, 0.98 and 0.91 for the three stages.

\subsubsection{Channel Idling}
The number of output and input channels usually does not correspond to the number of output and input channels processed in parallel by this core. The output and input channels are partitionned into blocks of $n_{ch} \times n_{ch}$ and filling in all-zero filters for the unused cases. The outputs of these blocks then have to be summed up pixel-wise off-chip. 

This processing in blocks can have a strong additional impact on the efficiency when not fully utilizing the core. While for the reasonable choice $n_{ch}=8$ the stages 2 and 3 of our reference ConvNet can be perfectly split into $n_{ch} \times n_{ch}$ blocks and thus no performance is lost, Stage~1 has only 3 input channels and can load the core only with $\eta_{blocks}=3/8$. However, stages with a small number of input and/or output channels generally perform much less operations and efficiency in these cases is thus not that important. 
 
The total throughput with the reference ConvNet running on this device with the configuration used for our implementation (cf. Section~\ref{sec:impl}) is summarized in Table~\ref{tbl:throughputEfficiency}, alongside details on the efficiency of the individual stages. 

\subsection{System Architecture}
\label{sec:sysArch}
When designing the architecture it is important to keep in mind how it can be used in a larger system. This system should be able to take a video stream from a camera, analyze the content of the images using ConvNets (scene labeling, object detection, recognition, tracking), display the result, and transmit alerts or data for further analysis over the network. 

\begin{figure}
	\includegraphics[width=\columnwidth]{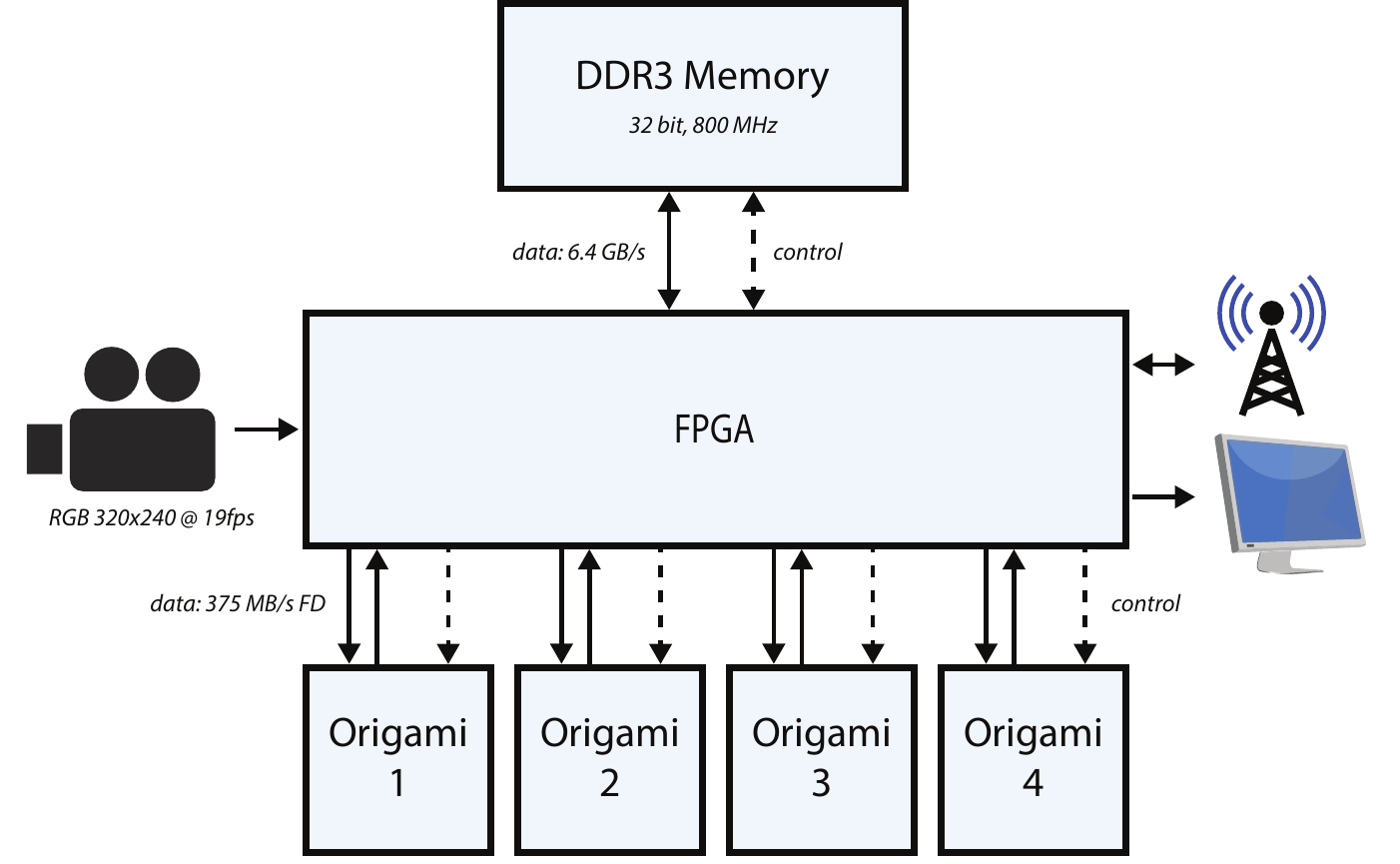}
	\caption{Suggested system architecture using dedicated Origami chips. The same system could also be integrated into a SoC.}
	\label{fig:systemArch}
\end{figure}

\subsubsection{General Architecture}
We elaborate one configuration (cf. Fig.~\ref{fig:systemArch}), based on which we show the advantages of our design. Besides the necessary peripherals and four Origami chips, there is a 32\,bit 800\,MHz DDR3 or LPDDR3 memory. The FPGA could be a Xilinx Zynq 7010 device\footnote{Favorable properties: low-cost, ARM core for control of the circuit and external interfaces, decent memory interface.}. 
The FPGA has to be configured to include a preprocessing core for rescaling, color space conversion, and local contrastive normalization. To store the data in memory after preprocessing, but also to load and store the data when applying the convolutional layer using the Origami chips and applying the fully-connected layers, there has to be a DMA controller and a memory controller. 

The remaining steps of the ConvNet like summing over the partial results returned by the Origami chips, adding the bias, and applying the ReLU non-linearity and max-pooling have to be done on the FPGA, but requires very little resources since no multipliers are required. The only multipliers are required to apply the fully-connected layers following the ConvNet feature extraction, but these do not have to run very fast, since the share of operations in this layer is less than 2\% for the scene labeling ConvNet in~\cite{Cavigelli2015}. 

For every stage of the ConvNet, we just tile the data into blocks of height $h_{in,max}$, $n_{ch}$ input channels and $n_{ch}$ output channels. We then sum up these blocks over the input channels and reassemble the final image in terms of output channels and horizontal stripes. 

\subsubsection{Bandwidth Considerations}
In the most naive setup, this means that we need to be able to provide memory accesses for the full I/O bandwidth of every connected Origami chip together. However, we also need to load the previous value of each output pixel because the results are only the partial sums and need to be added for the final result. In any case the ReLU operation and max-pooling can be done in a scan-line procedure right after computing the final values of the convolutional layers, requiring only a buffer of $(\ell-1) h_{in,max}/\ell$ values for $\ell\times\ell$ max-pooling since the max operation can be applied in vertical and horizontal direction independently (one direction can be done locally). 

However, this is far from optimal and we can improve using the same concept as in the Origami chip itself. We can arrange the Origami chips such that they calculate the result of a larger tile of input and output channels, making chips 1\&2 and 3\&4 share the same input data and chips 1\&3 and 2\&4 generate output data which can immediately be summed up before writing it to memory. Analogous to the same principle applied on-chip, this saves a factor of two for read and write access to the memory. Of course, the same limitations as explained in the previous section also apply at the system level. 

The pixel-wise fully-connected layers can be computed in a single pass, requiring the entire image to be only loaded once. For the scene labeling ConvNet we require $256\cdot 64+64\cdot 8\approx 17\text{k}$ parameters, which can readily be stored within the FPGA alongside 64 intermediate values during the computations. 

This system can also be integrated into a SoC for reduced system size and lower cost as well as improved energy efficiency. This makes the low memory bandwidth requirement the most important aspect of the system, being able to run with a narrow and moderate-bandwidth memory interface translates to lower cost in terms of packaging, and significantly higher energy efficiency (cf. Section~\ref{sec:discPowerEff} for a more detailed discussion of this on a per-chip basis).

\section{Implementation}
\label{sec:impl}
We first present the general implementation of the circuit. Thereafter, we present the results of a fixed-point analyses to determine the number format. We finalize this section by summarizing the implementation figures and by taking a look at implementation aspects of the entire system. 

\subsection{General Implementation}
\label{sec:generalImpl}
As discussed in Section~\ref{sec:arch}, we operate our design with two clocks originating from the same source, where one is running at twice the frequency of the other. The slower clock $f=250$\,MHz is driving the I/O and the SRAM, but also other elements which do not need to run very fast, such as the image and filter bank. The SoP units and channel summers are doing most of the computation and run at $f_{fast}=2f=500$\,MHz to achieve a high area efficiency. To achieve this frequency, the multipliers and the subsequent adder tree are pipelined. We have added two pipeline stages for each, the multipliers and the adder tree. 

\begin{figure}
	\centering
	\small
	\includegraphics[width=0.439\columnwidth]{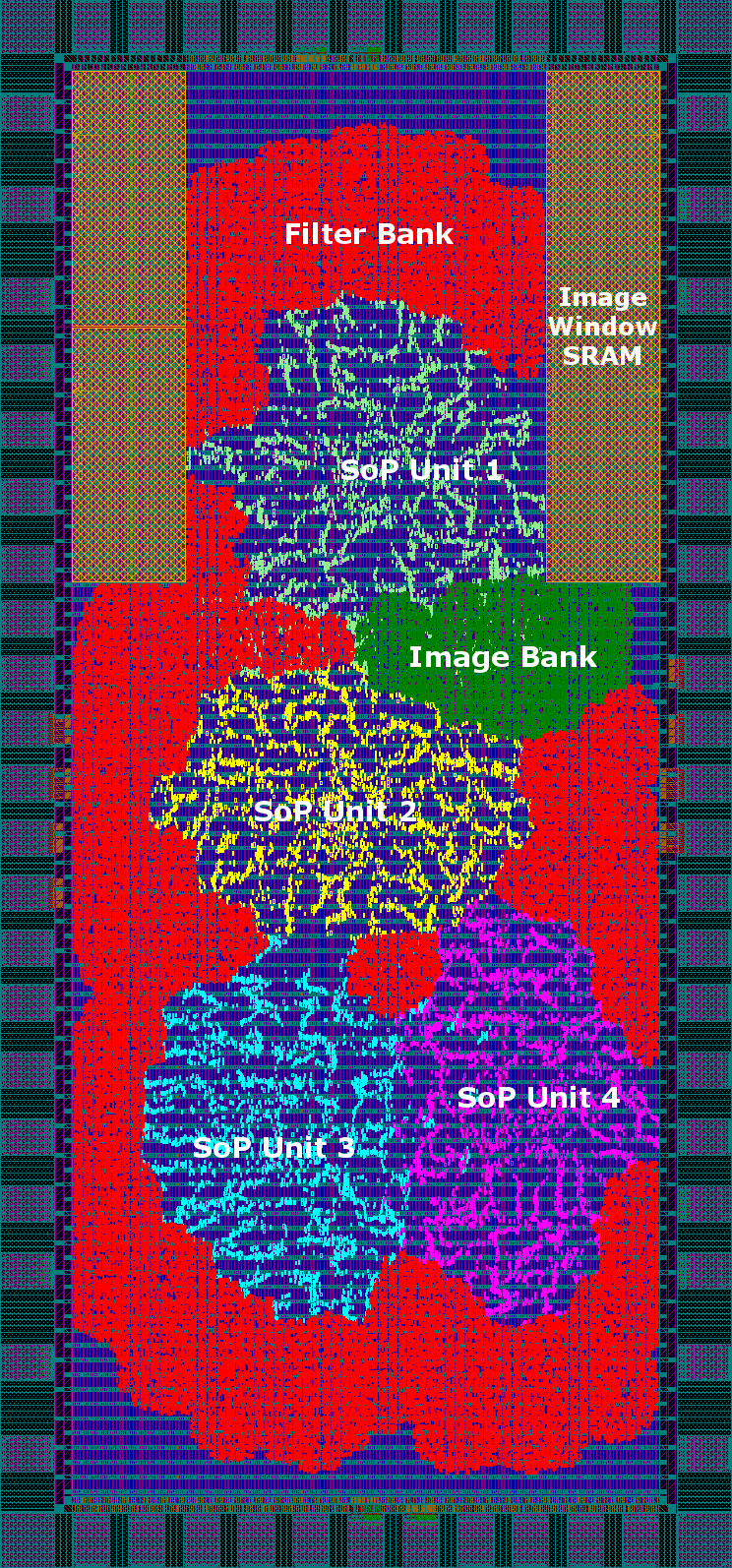}
	\includegraphics[width=0.456\columnwidth]{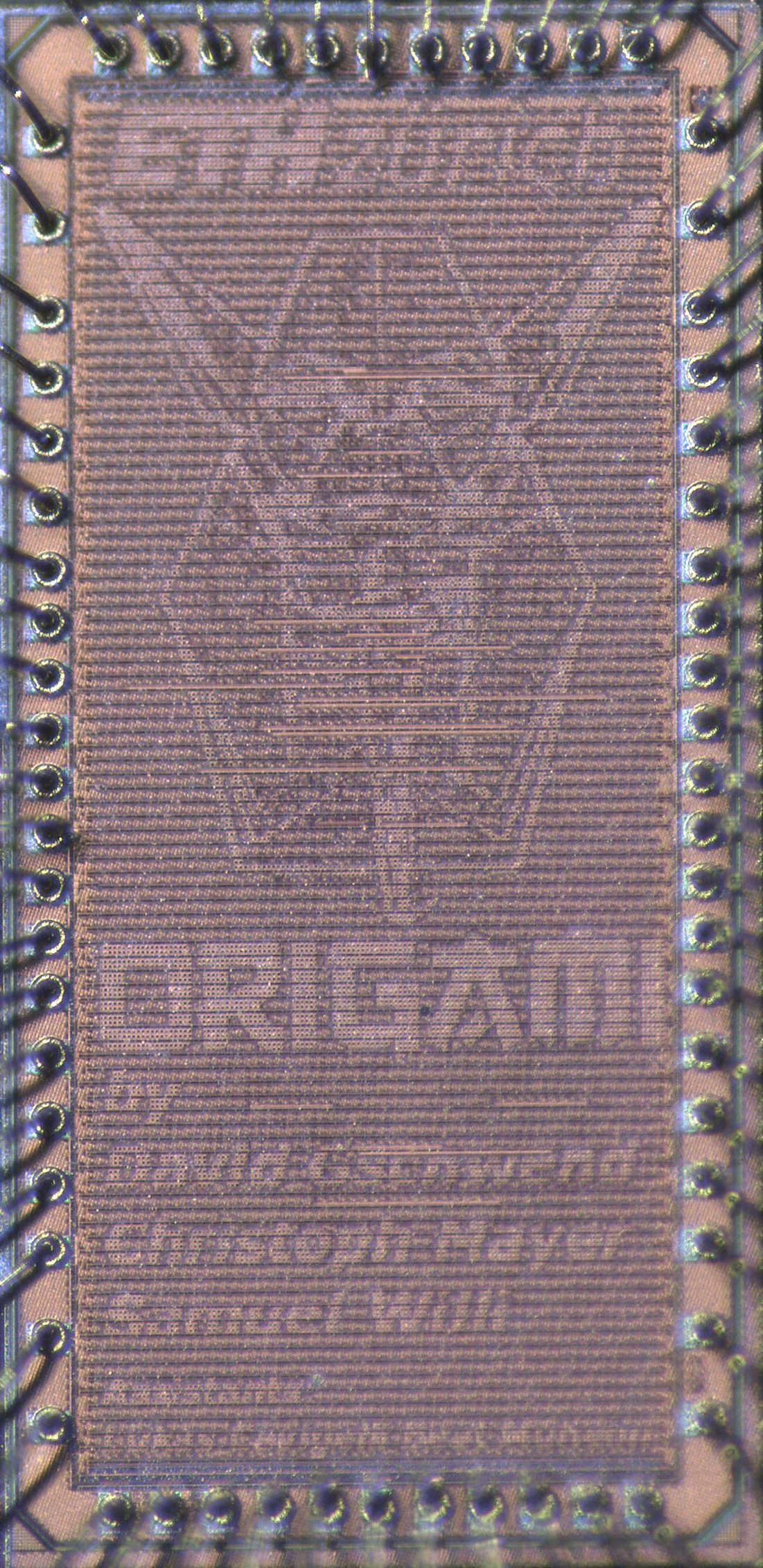}
	\caption{Floorplan and die shot of the final chip. In the floorplan view, the cells are colored by functional unit, making the low density in the sum-of-products computation units clearly visible.}
	\label{fig:dieshot}
\end{figure}

For the taped-out chip we set the filter size $h_k=w_k=7$, since we found $7\times 7$ and $5\times 5$ to be the most common filter sizes and a device capable of computing larger filter sizes is also capable to calculate smaller ones. For the maximum height of a horizontal image stripe we chose $h_{in,max}=512$, requiring an image window SRAM size of 29k words. Due to its size and latency, we have split it into four blocks of 1024 words each with a word width of $7\cdot 12$, as can be seen in the floorplan (Figure~\ref{fig:dieshot}). 
The alignment shown has resulted in the best performance of various configurations we have tried. Two of the RAM macro cells are placed besides each other on the left and the right boundary of the chip, with two of the cells flipped such that the ports face towards the center of the device. 
For silicon testing, we have included a built-in self-test for the memory blocks. 

The pads of the input data bus were placed at the top of the chip around the image bank memory, in which it is stored after one pipeline stage. The output data bus is located at the bottom-left together with an in-phase clock output and the test interface at the bottom-right of the die. The control and clock pads can be found around the center of the right side. Two Vdd and GND core pads were placed at the center of the left and right side each, and one Vdd and GND core pad was placed at the top and bottom of the chip. A pair of Vdd and GND pads for I/O power was placed close to each corner of the chip. 

The core clock of 500\,MHz is above the capabilities of standard CMOS pads and on-chip clock generation is unsuitable for such a small chip, while also complicating testing. To overcome this, two phase-shifted clocks of 250\,MHz are fed into the circuit. One of the clock directly drives the clock tree of the slower clock domain inside the chip. This clock is also XOR-ed with the second input clock signal to generate the faster 500\,MHz clock. 


\subsection{Fixed-Point Analysis}
\label{sec:fixedPoint}
Previous work is not conclusive on the required precision for ConvNets, 16 and 18\,bit are the most common values~ \cite{Farabet2009,Farabet2011,Gokhale2014,Conti}. To determine the optimal data width for our design, we performed a fixed-point analysis based on our reference ConvNet. We replaced all the convolution operations in our software model with fixed-point versions thereof and evaluated the resulting precision depending on the input, output and weight data width. The quality was analyzed based on the per-pixel classification accuracy of 150~test images omitted during training. We used the other 565~images of the Stanford backgrounds dataset~\cite{Gould2009} to train the network. 

\begin{figure}
	\centering	
	\includegraphics[width=0.6\columnwidth]{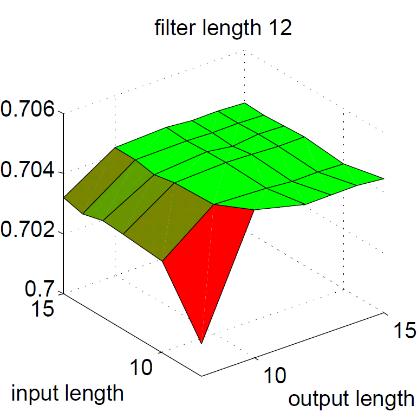}
	\caption{Classification accuracy with filter coefficients stored with 12\,bit precision. The single precision implementation achieves an accuracy of 70.3\%. Choosing an input length of 12\,bit results in an accuracy loss of less than 0.5\%.}
	\label{fig:fixedpoint}
\end{figure}

Our results have shown that an output length of 12\,bit is sufficient to keep the implementation loss below a drop of 0.5\% in accuracy. Since the convolution layers are applied repeatedly with little processing between them, we chose the same signal width for the input, although we could have reduced them further. For the filter weights a signal width of 12\,bit was selected as well.

For the implementation, we fixed the filter size $h_k=w_k=7$ and chose $n_ch=8$. Smaller filters have to be zero-padded and larger filters have to be decomposed into multiple $7 \times 7$ filters and added up. To keep the cycles lost during column changes low also for larger image, we chose $h_{(in,max)}=512$. For the SRAM, the technology libraries we used did not provide a fast enough module to accommodate $8 \cdot 512$~words of $7 \cdot 12$\,bit at 250\,MHz. Therefore, the SRAM was split into 4~modules of 1024~words each.


\subsection{System Implementation}

The four Origami chips require 750\,MB/s FD for the given implementation(12\,bit word length, $n_{ch}=8$ input and output channels, 250\,MHz) using the input and output feature map sharing discussed in Section~\ref{sec:sysArch} to save a factor of 2. The inputs and outputs are read directly from and again written directly to memory. Since the chips output partial sums, these have to be read again from memory and added up for the final convolution result. This can be combined with activation and pooling, adding slightly less than 750\,MB/s read and 187.5\,MB/s (for 2$\times$2 pooling) write memory bandwidth. For the third stage in the scene labeling ConvNet there is no such pooling layer, instead there is the subsequent pixel-wise classification which can be applied directly and which reduces the feature maps from 256 to 8, yielding an even lower memory write throughput requirement. To sum up, we require 2.45\,GB/s memory bandwidth during processing. To achieve the maximum performance, we have to load the filters at full speed for all four chip independently at the beginning of each processing burst, requiring a memory bandwidth of 1.5\,GB/s -- less than during computation. This leaves enough bandwidth available for some pre- and post-processing and memory access overhead. In the given configuration, $320\times 240$ frames can be processed at over 75 frame/s or at an accordingly higher resolution.

\section{Results}
\label{sec:results}

We have analyzed and measured multiple metrics in our architecture: I/O bandwidth, throughput per area and power efficiency. We can run our chip at two operating points: a high-speed configuration with $V_{dd}=1.2$\,V and a high-efficiency configuration with $V_{dd}=0.8$\,V. We have taped out this chip and the results below are actual silicon measurement results (cf. Table~\ref{tbl:postLayoutFigures}), as opposed to post-layout results which are known to be quite inaccurate. 
We now proceed by presenting implementation results and silicon measurements.

\subsubsection{Implementation}

\begin{figure}
	\centering
	\small
	\includegraphics[width=0.6\columnwidth]{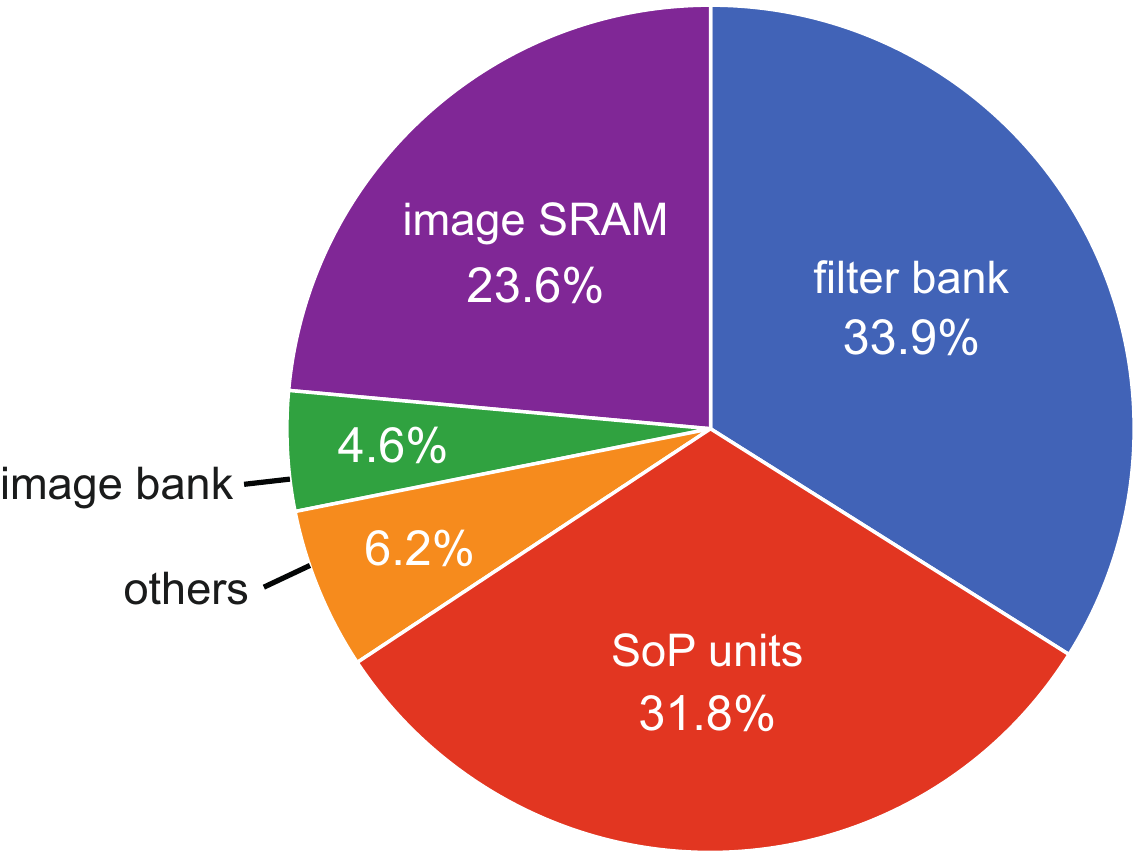}
	\caption{Final core area breakdown.}
	\label{fig:coreAreaBreakdown}
\end{figure}

We show the resulting final area breakdown in Figure~\ref{fig:coreAreaBreakdown}. 
The filter bank accounts for more than a third of the area and consists of registers storing the filter weights ($0.41\,mm^2$) and the multiplexers switching between them ($0.03\,mm^2$). The SoP units take up almost another third of the circuit and consist mostly of logic ($98188\,\mu m^2$/unit) and some pipeline registers ($26615\,\mu m^2$/unit). The rest of the space is shared between the image window SRAM, the image bank ($0.05\,mm^2$ registers and $7614\mu m^2$ logic), and other circuitry (I/O registers, data output bus mux, control logic; $0.1\,mm^2$). The chip area is clearly dominated by logic and is thus suited to benefit from voltage and technology scaling. 

We have used Synopsys Design Compiler 2013.12 for synthesis and Cadence SoC Encounter 13.14 for back-end design. Synthesis and back-end design have been performed for a clock frequency of $f=350$\,MHz with typical case corners for the functional setup view and best case corners for the functional hold view. Clock trees have been synthesized for the fast and the slow clock with a maximum delay of 240\,ps and a maximum skew of 50\,ps. For scan testing, a different view with looser constraints was used. Reset and scan enable signals have also been inserted as clock trees with relaxed constraints. Clock gating was not used. We performed a post-layout power estimation based on switching activity information from a simulation running parts of the scene labeling ConvNet. The total core power was estimated to be 620.8\,mW, of which 35.5\,mW are used in the $f_{fast}$ and 41.7\,mW are used in the lower frequency clock tree. Each SoP unit uses 66.9\,mW, the filter bank 122.5\,mW, the image bank 18.3\,mW, and the image window SRAM 43.6\,mW. The remaining power was used to power buffers connecting these blocks, I/O buffers and control logic. The entire core has only one power domain with a nominal voltage of 1.2\,V and the pad frame uses 1.8\,V. The power used for the pads is 205\,mW with line termination ($50\Omega$ towards 0.9\,V).


\subsubsection{Silicon Measurements}
The ASIC has been named \texttt{Origami} and has been taped-out in UMC 65nm CMOS technology. 
The key measurement results of the ASIC have been compiled in Table~\ref{tbl:postLayoutFigures}.  
In the high-speed configuration we can apply a 500\,MHz clock to the core, achieving a peak throughput of 196\,GOp/s. Running the scene labeling ConvNet from \cite{Cavigelli2015}, we achieve an actual throughput of 145\,GOp/s while the core (logic and on-chip memory) consumes 448\,mW. This amounts to a power efficiency of 437\,GOp/s/W, measured with respect to the peak throughput for comparability to related work. The I/O data interface consist of one input and one output 12\,bit data bus running at half of the core frequency, providing a peak bandwidth of 375\,MB/s full-duplex. We achieve a very high throughput density of $63.4\,\text{GOp/s/mm}^2$ despite the generously chosen core area of $3.09\,\text{mm}^2$ (to accommodate a large enough pad frame for all 55 pins), while the logic and on-chip memory occupy a total area of just $1.31\,\text{mm}^2$, which would correspond to a throughput density of $150\,\text{GOp/s/mm}^2$.  

\begin{table}
	\caption{Measured Silicon Key Figures.}
	\label{tbl:postLayoutFigures}
	\centering
	\small
	\begin{threeparttable}
	\begin{tabularx}{\columnwidth}{Xr}
		\toprule \multicolumn{2}{l}{Physical Characteristics} \\ \midrule
		Technology			& UMC 65\,nm, 8 Metal Layers \\
		Core/Pad Voltage	& 1.2\,V / 1.8\,V \\
		Package				& QFN-56 \\
		\# Pads				& 55 (i: 14, o: 13, clk/test: 8, pwr: 20) \\
		Core Area			& $3.09\,\text{mm}^2$ \\
		Circuit Complexity\tnote{a}	& 912\,kGE ($1.31\,\text{mm}^2$) \\
		Logic (std. cells)	& 697\,kGE ($1.00\,\text{mm}^2$) \\
		On-chip SRAM		& 344\,kbit \\ \midrule
		\multicolumn{2}{l}{Performance \& Efficiency @1.2\,V} \\ \midrule
		Max. Clock Frequency	& core: 500\,MHz, i/o: 250\,MHz \\
		Power\tnote{a} @500\,MHz	& 449\,mW (core) + 205\,mW (pads) \\ 
		Peak Throughput		& 196\,GOp/s \\
		Effective Throughput	& 145\,GOp/s \\
		Core Power-Efficiency	& 437\,GOp/s/W \\ \midrule
		\multicolumn{2}{l}{Performance \& Efficiency @0.8\,V} \\ \midrule
		Max. Clock Frequency	& core: 189\,MHz, i/o: 95\,MHz \\
		Power\tnote{b} @189\,MHz	& 93\,mW (core) + 144\,mW (pads) \\
		Peak Throughput		& 74\,GOp/s \\
		Effective Throughput	& 55\,GOp/s \\
		Core Power-Efficiency	& 803\,GOp/s/W \\
		\bottomrule
	\end{tabularx}
    \begin{tablenotes} 
    	\item[a]{Including the SRAM blocks.} 
    	\item[b]{The power usage was measured running with real data and at maximum load.} 
    \end{tablenotes} 
	\end{threeparttable}
\end{table}

\begin{figure}
	\centering
	\includegraphics[width=0.9\columnwidth]{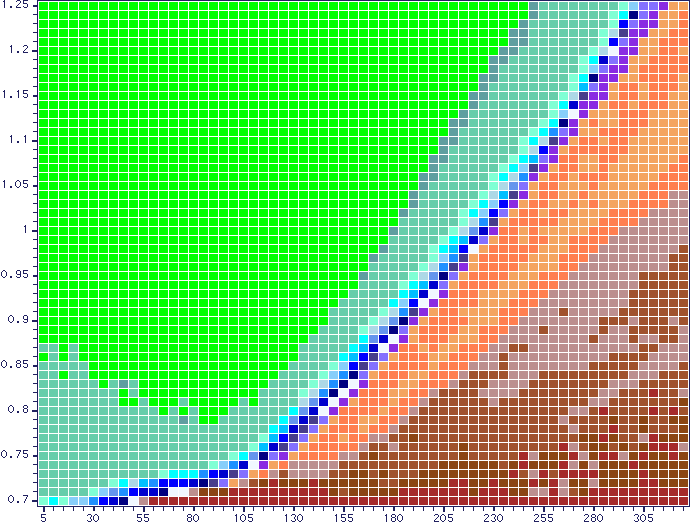}
	\caption{Shmoo plot showing number of incorrect results in dependence of frequency ($f=f_\text{fast}/2$, $x$-axis, in MHz) and core voltage ($V_\text{core}$, $y$-axis, in V) at 125°C. Green means no errors. }
	\label{fig:shmoo125}
\end{figure}

When operating our chip in the high-efficiency configuration, the maximum clock speed without inducing any errors is 189\,MHz. The throughput is scaled accordingly to 74\,GOp/s for the peak performance and 55\,GOp/s running our reference ConvNet. The core's power consumption is reduced dramatically to only 93\,mW, yielding a power-efficiency of 803\,GOp/s/W. The required I/O bandwidth shrinks to 142\,MB/s full-duplex or 1.92\,MB/GOp. The throughput density amount to $23.9\,\text{GOp/s/mm}^2$ for this configuration. The chip was originally targeted at the 1.2\,V operating point and has hold violations operating at 0.8\,V at room temperature. Thus the measurement have been obtained at a forced ambient temperature of 125°C. The resulting Shmoo plot is shown in Figure~\ref{fig:shmoo125}.   

Besides the two mentioned operating points there are many more, allowing for a continuous trade-off between throughput and energy efficiency by changing the core supply voltage as evaluated empirically in Figure~\ref{fig:energyEffThroughput}. As expected the figures are slightly worse for the measurements at higher temperature. Static power dissipation takes a share around 1.25\% across the entire voltage range at 25°C and a share of about 10.5\% in the interval $[0.95\,\text{V},1.25\,\text{V}]$ increasing to 14.7\% for a core voltage of 0.8\,V at 125°C.  

\begin{figure}
	\centering
	\begin{tikzpicture}[scale=0.75]
	\tikzset{mark options={solid, mark size=3, line width=1pt}} %
	\pgfplotsset{
 	 xmin=0.75, xmax=1.299
	}
	\ifarxiv	
		\newcommand{\effThrghptFile}{efficiencyThroughput.csv}
	\fi
	\begin{axis}[height=7cm, width=1.33\columnwidth-1.5cm, axis y line*=left, ymin=0, ymax=1050, xlabel=$V_\text{core}$ {[V]}, ylabel=Efficiency {[GOp/s/W]}, grid=major] 
		\addplot [thick, mark=diamond*, color=orange] table [x=vcore25, y=efficiency25, col sep=comma] 
			{\effThrghptFile}; \label{plotEffcyThrgpt:Eff25}
		\addplot [densely dashed, thick, mark=diamond*, color=green!50!black] table [x=vcore125, y=efficiency125, col sep=comma]
			{\effThrghptFile}; \label{plotEffcyThrgpt:Eff125}
	\end{axis}
	\begin{axis}[height=7cm, width=1.33\columnwidth-1.5cm, axis y line*=right, axis x line=none, ymin=0, ymax=263, ylabel=Throughput {[GOp/s]}]
		\addplot [thick, mark=x, color=blue] table [x=vcore25, y=throughput25, col sep=comma] 
			{\effThrghptFile}; \label{plotEffcyThrgpt:Thr25}
		\addplot [densely dashed, thick, mark=x, color=red] table [x=vcore125, y=throughput125, col sep=comma] 
			{\effThrghptFile}; \label{plotEffcyThrgpt:Thr125}
	\end{axis}
	\end{tikzpicture}\\
	\begin{scriptsize}	
		Efficiency 25°C (\ref{plotEffcyThrgpt:Eff25}), 125°C (\ref{plotEffcyThrgpt:Eff125}), Throughput 25°C (\ref{plotEffcyThrgpt:Thr25}), 125°C (\ref{plotEffcyThrgpt:Thr125})
	\end{scriptsize}
	\caption{Measured energy efficiency and throughput in dependence of $V_\text{core}$ for 25°C and 125°C.}
	\label{fig:energyEffThroughput}
\end{figure}

\section{Discussion}
\label{lbl:discussion}
None of the previous work on ConvNet accelerators has silicon measurement results. We will thus compare to post-layout and post-synthesis results of state-of-the-art related works, although such simulation results are known to be optimistic. We have listed the key figures of all these works in Table~\ref{tbl:comparison} and discuss the various results in the sections below. 

\begin{table*}
	\caption{Summary of Related Work for a Wide Range of Platforms (CPU, GPU, FPGA, ASIC).}
	\label{tbl:comparison}
	\setlength{\extrarowheight}{.75ex}
	\begin{threeparttable}
	\begin{tabularx}{\textwidth}{ll l p{8mm} p{8mm} p{8mm} p{13mm} p{10mm} l p{6mm} l l}
	\toprule
	publication & type & platform & theor.\tnote{a} & peak\tnote{a} & act.\tnote{a} & power\tnote{b} & power eff. & prec. & $V_\text{core}$ & area\tnote{j} & area eff.\tnote{h} \\
	 &  &  & GOp/s & GOp/s & GOp/s & W & GOp/s/W &  & V & MGE & GOp/s/MGE \\ \midrule
	Cavigelli et al.~\cite{Cavigelli2015} & CPU & Xeon \mbox{E5-1620v2} & 118 & & 35\tnote{d} & 230 & 0.15 & float32 & & & \\
	Cavigelli et al.~\cite{Cavigelli2015} & GPU & GTX780 & 3977 & 3030 & 1908\tnote{d} & sd:200 & 14\tnote{g} & float32 & & & \\ 
	cuDNN R3 \cite{ChintalaBenchm} & GPU & Titan X & 6600 & 6343 &  & d:250 & 25.6\tnote{g} & float32 & & & \\
	Cavigelli et al.~\cite{Cavigelli2015} & SoC & Tegra K1 & 365 & 95 & 84\tnote{d} & s:11 & 8.6 & float32 & & & \\
	\midrule
	CNP \cite{Farabet2009} & FPGA & Virtex4 & 40 & 40 & 37 & s:10 & 3.7 & fixed16 & & & \\
	NeuFlow \cite{Farabet2011} & FPGA & Virtex6 VLX240T & 160 & 160 & 147 & s:10 & 14.7 & fixed16 & & & \\
	nn-X \cite{Gokhale2014} & FPGA & Zynq XC7Z045 & 227 & 227 & 200 & s:8 d+m:4 & 25 & fixed16 & & & \\
	Zhang et al.~\cite{Zhang2015} & FPGA/HLS & Virtex7 VX485T & 62 & 62 &  & s:18.6 & 3.3 & float32 & & & \\
	\midrule
	ConvEngine \cite{Qadeer2013} & synth. & 45nm & 410 & & & c:1.0 & 409 & fixed10 & 0.9 & & \\
	ShiDianNao \cite{Du2015} & layout & TSMC 65nm & 128 & & & c:0.32 & 400 & fixed16 & & d\tnote{h}:4.86 & 26.3\\
	NeuFlow \cite{Farabet2011} & layout & IBM 45nm SOI & 1280 & 1280 & 1164 & d:5 & 230 & fixed16 & 1.0 & d:38.46 & 33.3\\
	NeuFlow \cite{Pham2012} & layout & IBM 45nm SOI & 320 & 320 & 294 & c:0.6 & 490 & fixed16 & 1.0 & d:19.23 & 16.6\\
	HWCE \cite{Conti} & layout & ST 28nm FDSOI & 37 & 37 & & c:0.18 & 206 & fixed16 & 0.8 & & \\
	HWCE \cite{Conti} & layout & ST 28nm FDSOI & 1 & 1 & & c:0.73m & 1375 & fixed16 & 0.4 & & \\ \midrule
	this work & silicon & umc 65nm & 196 & 196 & 145\tnote{d} & c:0.51\tnote{f} & 437 & fixed12 & 1.2 & c:0.91 d:2.16 & 90.7 \\
	this work & silicon & umc 65nm & 74 & 74 & 55\tnote{d} & c:0.093\tnote{f} & 803 & fixed12 & 0.8 & c:0.91 d:2.16 & 34.3 \\
	\bottomrule
	\end{tabularx}
    \begin{tablenotes} 
    	\item[a]{We distinguish between theoretical performance, where we consider the maximum throughput of the arithmetic units, the peak throughput, which is the maximum throughput for convolutional layers of any size, and the actual throughput, which has been benchmarked for a real ConvNet and without processing in batches.} 
    	\item[b]{For the different types of power measurements, we abbreviate: s (entire system), d (device/chip), c (core), m (memory), io (pads), sd (system differential load-vs-idle).}
    	\item[c]{We use the abbreviations c (core area, incl. SRAM), d (die size)}
    	\item[d]{These values were obtained for the ConvNet described in~\cite{Cavigelli2015}.}
    	\item[f]{The static power makes up for around 1.3\% of the total power at 25°C for the entire range of feasible $V_\text{core}$, and about 11\% at 125°C.}
    	\item[g]{The increased energy efficiency of the Titan~X over the GTX780 is significant and can neither be attributed solely to technology (both 28\,nm) nor the software implementation or memory interface (both GDDR5). Instead the figures published by Nvidia suggest that architectural changes from Kepler to Maxwell are the source of this improvement.}
    	\item[h]{We take the theoretical performance to be able to compare more works and the device/chip size for the area. ShiDianNao does not include a pad ring in their layout (3.09$\text{mm}^2$), so we added it for better comparability (height 90$\mu$m).}
    	\item[j]{We measure area in terms of size of millions of 2-input NAND gates. 1GE: 1.44$\mu \text{m}^2$ (umc 65nm), 1.17$\mu \text{m}^2$ (TSMC 65nm), 0.65$\mu \text{m}^2$ (45nm), 0.49$\mu\text{m}^2$ (ST 28nm FDSOI).}
    \end{tablenotes} 
	\end{threeparttable}
\end{table*}

\subsection{Area Efficiency}
Our chip is the most area-efficient ConvNet accelerator reported in literature. We measure the area in terms of 2-input NAND gate equivalents to compensate for technology differences to some extent. With 90.7\,GOp/s/MGE our implementation is by far the most area-efficient, and even in high power-efficiency configuration we outperform previous state-of-the-art results. 
The next best implementation is a NeuFlow design at 33.8\,GOp/s/MGE, requiring a factor 3 more space for the same performance. 
ShiDianNao is of comparable area-efficiency with 26.3\,GOp/s/MGE. 
Also note that the chip size was limited by the pad-frame, and that the area occupied by the standard cells and the on-chip SRAM is only $1.31\,\text{mm}^2$ (0.91\,MGE). We would thus achieve a throughput density of an enormous $215\,\text{GOp/s/MGE}$ in this very optimistic scenario. This would require a more complex and expensive pad-frame architecture, e.g. flip-chip with multiple rows of pads, which we decided not to implement. 

We see the reason for these good results in our approach to compute multiple input and output channels in parallel. This way we have to buffer a window of 8 input channels to compute 64 convolutions, instead of buffering 64 input images, a significant saving of storage, particularly also because the window size of the input images that has to be buffered is a lot larger than the size of a single convolution kernel. Another ~25\% can be attributed to the use of 12\,bit instead of 16\,bit words which expresses itself mostly with the size of the SRAM and the filter kernel buffer.  

\subsection{Bandwidth Efficiency}
The number of I/O pins is often one of the most congested resources when designing a chip, and the fight for bandwidth is even more present when the valuable memory bandwidth of an SoC has to be shared with accelerators. We achieve a bandwidth efficiency of 521\,GOp/GB, providing an improvement by a factor of more than $10\times$ over the best previous work -- NeuFlow comes with a memory bandwidth of 6.4\,GB/s to provide 320\,GOp/s, i.e. it can perform 50\,GOp/GB. ShiDianNao does not provide any information on the external bandwidth and the HWCE can do 6.1\,GOp/GB. These large differences, particularly between this work and previous results, can be attributed to us having focused on reducing the required bandwidth while maximizing throughput on a small piece of silicon. The architecture has been designed to maximize reuse of the input data by calculating pixels of multiple output channels in parallel, bringing a significant improvement over caching as in \cite{Du2015} or accelerating individual 2D convolutions \cite{Qadeer2013,Conti}. 

\subsection{Power Efficiency}
\label{sec:discPowerEff}
Our chip performs second-best in terms of energy efficiency of the core with 803\,GOp/s/W (high-efficiency configuration) and 437\,GOp/s (high-performance configuration), being outperformed only by the HWCE. The HWCE can reach up to 1375\,GOp/s/W in its high-efficiency setup when running at 0.4\,V and making use of reverse body biasing, available only with FDSOI technology to this extent. Our chip is then followed by NeuFlow (490\,GOp/s/W), the Convolution Engine (409\,GOp/s/W) and ShiDianNao (400\,GOp/s/W).  

However, technology has a strong impact on the energy efficiency. Our design was done in UMC~65\,nm, while NeuFlow was using IBM~45\,nm SOI and HWCE even resorted to ST~28\,nm~FDSOI. In order to analyze our architecture independently of the particular implementation technology used, we take a look at its effect. We take the simple model
\begin{align*}
\tilde{P}=P \frac{\ell_{new}}{\ell_{old}} \left(\frac{V_{dd,new}}{V_{dd,old}}\right)^2. 
\end{align*}
The projected results are shown in Table~\ref{tbl:scaledPowerEff}. To obtain the operating voltage in 28\,nm technology, we scale the operating voltage linearly with respect to the common operating voltage of the used technology. This projection, although not based on a very accurate model, gives an idea of how the various implementations perform in a recent technology. Clearly, the only competitive results in terms of core power efficiency are ShiDianNao and this work.  
\begin{table}
	\caption{Projected Power and Power-Efficiency When Scaled to 28\,nm Technology}
	\label{tbl:scaledPowerEff}
	\centering
	\begin{tabular}{lrrr}
	\toprule
		publication & $V_{core}$ & power & efficiency\\
		 & V & mW & GOp/s/W \\ \midrule
		ConvEngine \cite{Qadeer2013} & 0.72 & 398 & 1030 \\
		ShiDianNao \cite{Du2015} & 0.8 & 61.3 & 2098 \\
		NeuFlow \cite{Farabet2011} & 0.8 & 239 & 1339 \\
		HWCE \cite{Conti} & 0.8 & 180 & 260 \\
		HWCE \cite{Conti} & 0.4 & 0.73 & 1375 \\
		this work & 0.8 & 86.1 & 2276 \\
		this work & 0.53 & 7.81 & 9475 \\ \bottomrule
	\end{tabular}
\end{table}
Previous work has always excluded I/O power, although it is a major contributor to the overall power usage. We estimate this power based on an energy usage of 21\,pJ/bit, which has been reported for a LPDDR3 memory model and the PHY on the chip in 28\,nm \cite{Schaffner2015}, assuming a reasonable output load and a very high page hit rate. For our chip, this amounts to an additional 63\,mW or 24\,mW for the high-performance and high-throughput configuration, respectively. For NeuFlow, due to the much higher I/O bandwidth, it looks worse with an additional 1.08\,W for their 320\,GOp/s implementation. 
If we assume the power efficiency of these devices in their original technology, this reduces the power efficiency including I/O to 342\,GOp/s/W, 632\,GOp/s/W and 191\,GOp/s/W for our chip in high-throughput and high-efficiency configuration as well as NeuFlow. If we look at their projected 28\,nm efficiency, they are decreased to 1315\,GOp/s/W, 2326\,GOp/s/W and 243\,GOp/s/W. This clearly shows the importance of the reduced I/O bandwidth in our design, and the relevance of I/O power in general with it making up a share of 42\% to 82\% of the total power consumption for these three devices.  

\section{Conclusions \& Future Work}
\label{sec:conclusions}

We have presented the first silicon measurement results of a convolutional network accelerator. The developed architecture is also first to scale to multi-TOp/s performance by significantly improving on the external memory bottleneck of previous architectures. It is more area efficient than any previously reported results and comes with the lowest-ever reported power consumption when compensating for technology scaling.

Further work with newer technologies, programmable logic and further configurability to build an entire high-performance low-power system is planned alongside investigations into the ConvNet learning-phase to adapt networks for very-low precision accelerators during training.  

\ifCLASSOPTIONcaptionsoff
  \newpage
\fi



\bibliographystyle{IEEEtran}
\bibliography{IEEEabrv,../library,refformat}
%
%
%

%
\vspace{-1.5cm}
\begin{IEEEbiography}[{\includegraphics[width=1in,height=1.25in,clip,keepaspectratio]{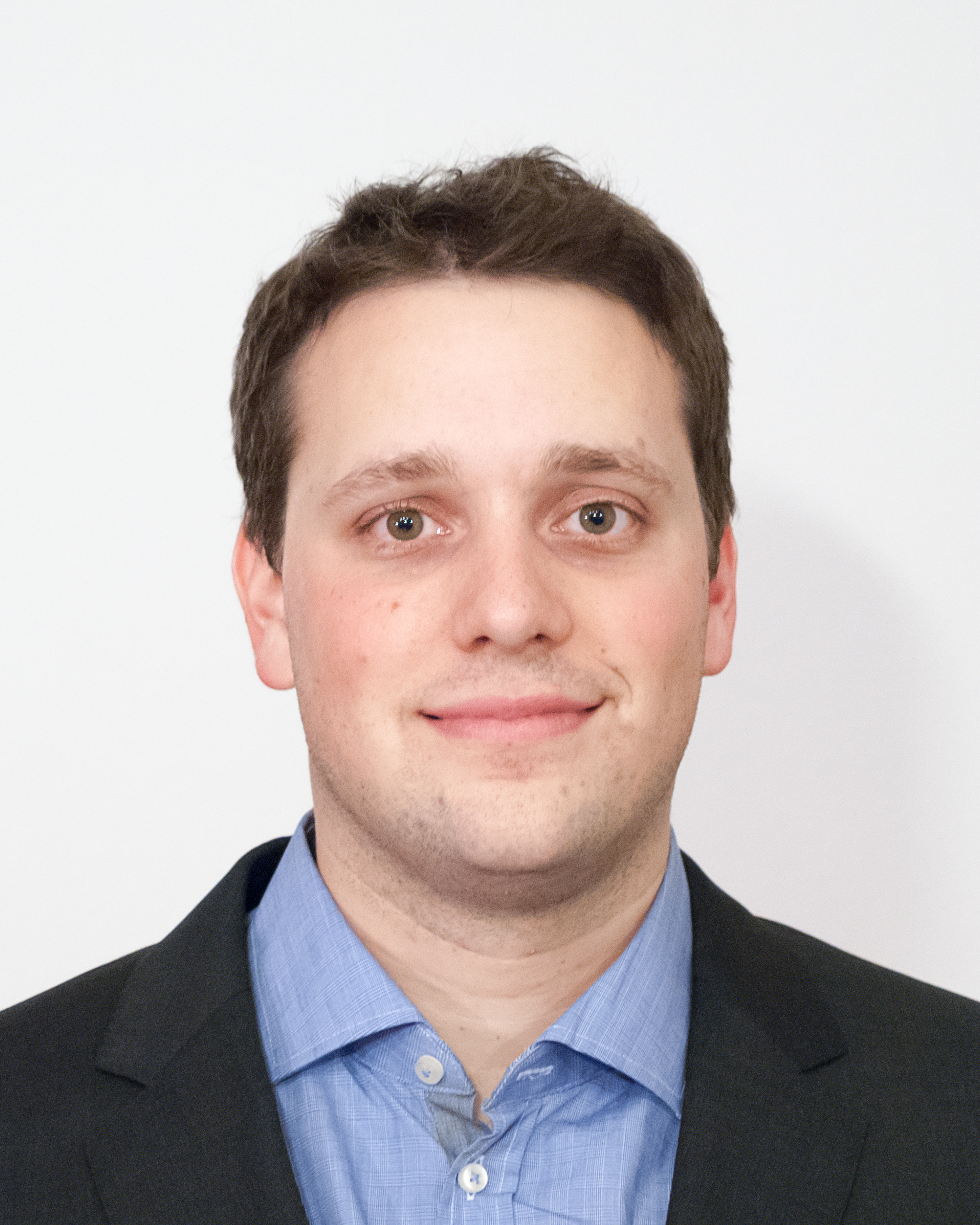}}]{Lukas Cavigelli} received the M.Sc. degree in electrical engineering and information technology from ETH Zurich, Zurich, Switzerland, in 2014.
Since then he has been with the Integrated Systems Laboratory, ETH Zurich, pursuing a Ph.D. degree. His current research interests include deep learning, computer vision, digital signal processing, and low-power integrated circuit design.
Mr. Cavigelli received the best paper award at the 2013 IEEE VLSI-SoC Conference.
\vspace{-1.5cm}
\end{IEEEbiography}
\begin{IEEEbiography}[{\includegraphics[width=1in,height=1.2in,clip,keepaspectratio]{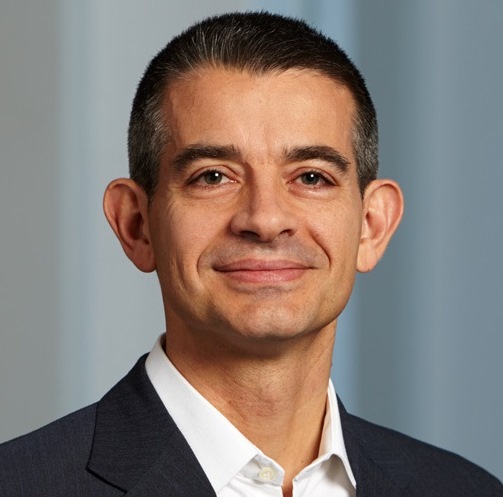}}]{Luca Benini} is the Chair of Digital Circuits and Systems at ETH Zurich and a Full Professor at the University of Bologna. He has served as Chief Architect for the Platform2012/STHORM project at STmicroelectronics, Grenoble. He has held visiting and consulting researcher positions at EPFL, IMEC, Hewlett-Packard Laboratories, Stanford University. Dr. Benini's research interests are in energy-efficient system and multi-core SoC design.  He is also active in the area of energy-efficient smart sensors and sensor networks for biomedical and ambient intelligence applications. 
He has published more than 700 papers in peer-reviewed international journals and conferences, four books and several book chapters. He is a Fellow of the IEEE and a member of the Academia Europaea. 
\vspace{-1.5cm}
\end{IEEEbiography}


 




\end{document}

%% file: timeline-v03.pdf_tex
\begingroup%
  \makeatletter%
  \providecommand\color[2][]{%
    \errmessage{(Inkscape) Color is used for the text in Inkscape, but the package 'color.sty' is not loaded}%
    \renewcommand\color[2][]{}%
  }%
  \providecommand\transparent[1]{%
    \errmessage{(Inkscape) Transparency is used (non-zero) for the text in Inkscape, but the package 'transparent.sty' is not loaded}%
    \renewcommand\transparent[1]{}%
  }%
  \providecommand\rotatebox[2]{#2}%
  \ifx\svgwidth\undefined%
    \setlength{\unitlength}{211.97260742bp}%
    \ifx\svgscale\undefined%
      \relax%
    \else%
      \setlength{\unitlength}{\unitlength * \real{\svgscale}}%
    \fi%
  \else%
    \setlength{\unitlength}{\svgwidth}%
  \fi%
  \global\let\svgwidth\undefined%
  \global\let\svgscale\undefined%
  \makeatother%
  \begin{picture}(1,0.57594414)%
    \put(0,0){\includegraphics[width=\unitlength,page=1]{timeline-v03.pdf}}%
    \put(0.30202541,0.38495774){\color[rgb]{0,0,0}\rotatebox{58.32624897}{\makebox(0,0)[lb]{\smash{load col $1$}}}}%
    \put(0.21703096,0.38495775){\color[rgb]{0,0,0}\rotatebox{58.32624897}{\makebox(0,0)[lb]{\smash{load weights}}}}%
    \put(0.35975442,0.38495774){\color[rgb]{0,0,0}\rotatebox{58.32624897}{\makebox(0,0)[lb]{\smash{load col $2$}}}}%
    \put(0.47326175,0.38495774){\color[rgb]{0,0,0}\rotatebox{58.32624897}{\makebox(0,0)[lb]{\smash{load col $w_k-1$}}}}%
    \put(0.52849874,0.38495774){\color[rgb]{0,0,0}\rotatebox{58.32624897}{\makebox(0,0)[lb]{\smash{load col $w_k$}}}}%
    \put(0.58501497,0.38495774){\color[rgb]{0,0,0}\rotatebox{58.32624897}{\makebox(0,0)[lb]{\smash{load col $w_k+1$}}}}%
    \put(0.69834333,0.38495774){\color[rgb]{0,0,0}\rotatebox{58.32624897}{\makebox(0,0)[lb]{\smash{load col $w_{in}$}}}}%
    \put(0,0){\includegraphics[width=\unitlength,page=2]{timeline-v03.pdf}}%
    \put(0.78360784,0.38495774){\color[rgb]{0,0,0}\rotatebox{58.32624897}{\makebox(0,0)[lb]{\smash{load weights}}}}%
    \put(0.86976704,0.38495775){\color[rgb]{0,0,0}\rotatebox{58.32624897}{\makebox(0,0)[lb]{\smash{load col $1$}}}}%
    \put(0.12324936,0.34017155){\color[rgb]{0,0,0}\makebox(0,0)[rb]{\smash{input data}}}%
    \put(0.12324936,0.25324962){\color[rgb]{0,0,0}\makebox(0,0)[rb]{\smash{output data}}}%
    \put(0.96985165,0.31813118){\color[rgb]{0,0,0}\makebox(0,0)[lb]{\smash{$t$}}}%
    \put(0,0){\includegraphics[width=\unitlength,page=3]{timeline-v03.pdf}}%
    \put(0.96985165,0.2426497){\color[rgb]{0,0,0}\makebox(0,0)[lb]{\smash{$t$}}}%
    \put(0,0){\includegraphics[width=\unitlength,page=4]{timeline-v03.pdf}}%
    \put(0.12324936,0.15146789){\color[rgb]{0,0,0}\makebox(0,0)[rb]{\smash{input data}}}%
    \put(0.12324936,0.06454599){\color[rgb]{0,0,0}\makebox(0,0)[rb]{\smash{output data}}}%
    \put(0.96985165,0.12942755){\color[rgb]{0,0,0}\makebox(0,0)[lb]{\smash{$t$}}}%
    \put(0,0){\includegraphics[width=\unitlength,page=5]{timeline-v03.pdf}}%
    \put(0.96985165,0.05394604){\color[rgb]{0,0,0}\makebox(0,0)[lb]{\smash{$t$}}}%
    \put(0,0){\includegraphics[width=\unitlength,page=6]{timeline-v03.pdf}}%
    \put(0.25763683,0.00313197){\color[rgb]{0,0,0}\makebox(0,0)[b]{\smash{$n_{ch}(h_k - 1)$}}}%
    \put(0,0){\includegraphics[width=\unitlength,page=7]{timeline-v03.pdf}}%
    \put(0.69805761,0.00270893){\color[rgb]{0,0,0}\makebox(0,0)[b]{\smash{$7+n_{ch}$}}}%
    \put(0.68924622,0.00050504){\color[rgb]{0,0,0}\makebox(0,0)[lb]{\smash{}}}%
    \put(-0.14104974,0.43452336){\color[rgb]{0,0,0}\makebox(0,0)[lt]{\begin{minipage}{0.28305544\unitlength}\raggedright \end{minipage}}}%
    \put(0.12313534,-0.15045786){\color[rgb]{0,0,0}\makebox(0,0)[lb]{\smash{}}}%
    \put(0,0){\includegraphics[width=\unitlength,page=8]{timeline-v03.pdf}}%
    \put(0.48408118,0.00313197){\color[rgb]{0,0,0}\makebox(0,0)[b]{\smash{out col $1$}}}%
    \put(0.84261813,0.00313197){\color[rgb]{0,0,0}\makebox(0,0)[b]{\smash{out col $2$}}}%
  \end{picture}%
\endgroup%